\documentclass[10pt]{article} 

\usepackage{hyperref}
\usepackage{url}
\usepackage{subcaption}
\usepackage{graphicx}
\usepackage{booktabs}
\usepackage[table]{xcolor} 
\usepackage{diagbox}
\usepackage{multirow}  
\usepackage{float}
\usepackage[most]{tcolorbox}
\usepackage{wrapfig}

\usepackage[accepted]{tmlr}

\usepackage{amsmath,amsfonts,bm}

\def\eqref#1{equation~\ref{#1}}

\def\1{\bm{1}}

\DeclareMathAlphabet{\mathsfit}{\encodingdefault}{\sfdefault}{m}{sl}
\SetMathAlphabet{\mathsfit}{bold}{\encodingdefault}{\sfdefault}{bx}{n}

\title{ARC-Encoder: learning compressed text representations for large language models}

\author{\name Hippolyte Pilchen \email hippolyte.pilchen@kyutai.org  
    \\
      \addr Kyutai, Paris, France
    \AND 
        \name Edouard Grave 
            \\ 
      \addr Kyutai, Paris, France
      \AND
      \name Patrick Perez 
      \\
      \addr Kyutai, Paris, France }

\def\month{07}  
\def\year{2026} 
\def\openreview{\url{https://openreview.net/forum?id=lU1P9dsqfn}} 

\definecolor{darkblue}{rgb}{0, 0, 0.5}
\hypersetup{colorlinks=true, citecolor=darkblue, linkcolor=darkblue, urlcolor=darkblue}

\newcommand{\myparag}[1]{\smallskip\noindent\textbf{#1}\;}

\newcommand{\modelname}[1]{ARC#1-Encoder}
\newcommand{\tightcmidrule}[1]{%
  \cmidrule(l{0pt}){#1}
}

\begin{document}

\maketitle

\begin{abstract}

Recent techniques such as retrieval-augmented generation or chain-of-thought reasoning have led to longer contexts and increased inference costs. Context compression techniques can reduce these costs, but the most effective approaches require fine-tuning the target model or even modifying its architecture. This can degrade its general abilities when not used for this specific purpose. Here we explore an alternative approach: an encoder that compresses the context into continuous representations which replace token embeddings in decoder LLMs. First, we perform a study of training strategies and architecture choices for the encoder. Our findings led to the design of an Adaptable text Representations Compressor, named ARC-Encoder, which outputs $x$ times fewer continuous representations (typically $x\!\in\!\{4,8\}$) than text tokens. We evaluate ARC-Encoder across a variety of LLM usage scenarios, ranging from in-context learning to context window extension, on both instruct and base decoders. Results show that ARC-Encoder achieves strong performance on several benchmarks and tasks while improving computational efficiency at inference. Finally, we demonstrate that our models can be adapted to multiple decoders simultaneously, allowing a single encoder to generalize across different decoder LLMs. This makes ARC-Encoder a flexible and efficient solution for portable encoders that can support multiple LLMs with only small MLPs.

\end{abstract}
\section{Introduction}

As their use expands, LLMs are required to process increasingly long contexts to incorporate detailed user prompts or external knowledge retrieved from large corpora as in retrieval-augmented generation (RAG) systems \citep{lewis2021retrievalaugmentedgenerationknowledgeintensivenlp}. However, this implies a \emph{large computational cost} at inference due to the quadratic complexity of Transformer attention mechanisms \citep{tay2022efficienttransformerssurvey}. Furthermore, using longer contexts can dilute interesting information leading to poor downstream results or can even reach the \emph{context window limit} of the LLM, which damages the model capacities. 

To address this issue, context compression is a promising solution. It reduces input length while preserving the semantics necessary for accurate generation \citep{liu2025shiftingaiefficiencymodelcentric}. Techniques fall into two main categories: \emph{hard compression}, which prunes, summarizes or deletes tokens \citep{jiang2023llmlinguacompressingpromptsaccelerated}, offering interpretability and model-agnosticism but limited compression; and \emph{soft compression}, which encodes context into dense vectors, e.g. memory tokens and gist tokens \citep{mu2024learningcompresspromptsgist, ge2024incontextautoencodercontextcompression}. While achieving higher compression, most existing soft methods require adapting the target decoder itself to process these representations, and necessitate training a completely separate, dedicated encoder for each individual decoder \citep{louis2025piscoprettysimplecompression,tang2025gmsaenhancingcontextcompression}.

Our work aims at leveraging LLMs' compression ability in cases
where documents should be processed on the fly. Most importantly, we constrain ourselves not to modify the decoder to preserve the LLM’s original generative capabilities while maintaining flexibility between compression and accuracy on downstream tasks.  We introduce an Adaptable text Representations Compressor (``\modelname{}'') that produces \textit{pooled tokens}, optimized to be directly consumable by a decoder in the same way as standard input tokens (injected after the embedding matrix). A single compressor module can be efficiently trained to support a predefined set of LLMs, ensuring cross-model compatibility. By preserving the few-shot abilities of models \citep{brown2020languagemodelsfewshotlearners}, we ensure compatibility with the question-answering (QA) evaluation set-up, such as 5-shot evaluation using exact match.
We achieve nearly the same accuracy as a decoder operating on full text, even at a $4 \times$ pooling factor while using an encoder adaptable to two decoders by means of 
a small MLP which has less than $1\%$ of the encoder parameters. 
Our approach achieves strong results among models that do not fine-tune the decoder while generalizing well to various tasks and domains. While methods that fine-tune the decoder may offer performance gains on the task specifically trained on as in standard supervised fine-tuning, our method eliminates the risk of forgetting and enables generalization across decoders. 
Our main contributions:

\begin{itemize}
\item We introduce \modelname{}, a method to compute compressed text representations that replace the raw text input in LLMs. Our approach reduces the input sequence length, \emph{without requiring any modification to the decoder model}. \modelname{} preserves strong performance across various benchmarks and scenarios, including in-context learning (ICL).
\item We show that a single encoder can be trained to work with multiple decoders, requiring less than $1\%$ new parameters per LLM. \modelname{} can be adapted to new decoders with minimal adjustment.
\item \modelname{} can be trained to extend a decoder context size, by compressing chunks of a large document in parallel, showing competitive results on long-context benchmarks.
\item Finally, we show that both pretraining and fine-tuning are key to the success of our approach. We also find that the compressed representations of Wikipedia require memory on the same order as that needed to store the raw text, allowing to precompute representations.
\end{itemize}

\section{Related work}

\paragraph{Encoder-Decoder architectures.}
Text auto-encoders have long been studied to improve transformers on specific downstream tasks. They process the input into dense embeddings to reduce processing cost while preserving or improving model accuracy. For instance, Atlas \citep{izacard2022atlasfewshotlearningretrieval} retrieves and encodes multiple relevant passages before decoding with a focus on knowledge intensive tasks. RAVEN \citep{huang2024ravenincontextlearningretrievalaugmented} uses a similar retrieval-augmented encoder-decoder structure improving in-context learning abilities while using less compute. More recently, \citet{t5gemma} propose an asymmetric architecture, closer to ours, where a smaller encoder aims at improving the decoder generation through cross-attention under similar inference budget. As in our work, they experimented with combination of different-sized models. 

\myparag{Context Compression.} Context compression reduces the number of tokens processed by a model to improve the efficiency of inference. There are two main approaches. 
\emph{Hard compression} methods, such as LLMLingua \citep{jiang2023llmlinguacompressingpromptsaccelerated}, operate directly in the text space by removing tokens from prompts. It aims at reducing their length while preserving the performance of the model. In contrast, we perform \emph{soft compression} which involves learning continuous compressed representations. This line of work began with gist tokens \citep{mu2024learningcompresspromptsgist}, which summarize task instructions into a few tokens by modifying the attention matrix to force generated tokens to only attend to gist tokens.
Similarly, in memory tokens \citep{chevalier2023adaptinglanguagemodelscompress,ge2024incontextautoencodercontextcompression},
learnable vectors which come from an encoder are prepended to the input sequence. These vectors serve as condensed representations of the full sequence when passed through the decoder. These methods typically rely on a pretraining phase to align the encoder’s output with the decoder’s hidden states, followed by fine-tuning of both encoder and decoder to fully leverage the compressed representations \citep{louis2025piscoprettysimplecompression,louis2025oscaronlinesoftcompression}. Recently, more similar to our pooling method, \citet{tang2025gmsaenhancingcontextcompression} explore using merged tokens to replace memory ones. They perform several training stages on the encoder, but also on the decoder in contrast to our work. Fine-tuning the decoder often degrades performance on standard tokens compared to our method which enables to use compressed tokens as well as standard ones. Other approaches explore the use of pre-computed text embeddings as memory tokens, reaching higher pooling factors (up to $\times 150$) with xRAG \citep{cheng2024xragextremecontextcompression} but performing poorly on certain benchmarks and lacking compression flexibility. All these context compression methods rely on the intrinsic compression capacity of LLMs. Indeed, \citet{kuratov2025cramming1568tokenssingle} has proven that an LLM decoder can be used to directly compress a text passage of roughly 1568 tokens into just one 4096-dimensional vector. More recently, \citet{eyuboglu2025cartridgeslightweightgeneralpurposelong} leveraged this property to produce compressed KV-caches for frequently used long documents. Alternative methods use vision encoders \citep{xing2025visioncentrictokencompressionlarge} to produce compressed textual representations from image-rendered text.

\myparag{Long Context.}
Recent work on long-context language modeling combines fine-tuning with extended positional encoding strategies. \citet{together2023Llama2_7b_32k_instruct} fine-tunes Llama2 7B to handle 32k-token inputs using position interpolation \citep{chen2023extendingcontextwindowlarge}. \citet{zhang2024longcontextcompressionactivation} inserts activation “anchors” into the hidden states of the model, requiring modification and fine-tuning of the target LLM itself to compress. In contrast, \citet{yen2024longcontextlanguagemodelingparallel} introduce a lightweight encoder that processes long inputs in parallel and passes compressed representations to a decoder via learned cross-attention, allowing efficient long-sequence handling. Our work follows the idea of \citet{yen2024longcontextlanguagemodelingparallel} while producing fewer tokens and keeping the decoder untouched.

\section{Method}
\label{sec:method}

\subsection{Architecture}

The overall architecture that we consider comprises a text \emph{encoder} and an \emph{MLP projector}, together forming a trainable \modelname{}, followed by the frozen target \emph{decoder}. The encoder is based on an LLM transformer, from which we remove the output head and the causal mask as usually done for sentence embedders \citep{lee2025nvembedimprovedtechniquestraining}. We add a pooling mechanism that reduces the number of elements in the sequence from $n$ to $\frac{n}{x}$ with $x$ the pooling factor (PF). The MLP projector is a 2-layer MLP without activation, mapping the encoder output to the embedding dimension of the decoder through a dimensional bottleneck. The decoder remains unchanged, as opposed to the encoder and MLP that are trained. The compressed continuous representations from the \modelname{} are used instead of the token embeddings in the decoder.

\subsection{Pooling method}
\label{subsec:pooling_method}
Most context compression works use learned or memory tokens as compressed representations. 
This makes it harder to effectively compress sequences of various lengths, as the number of output representations is fixed. Instead, we pool hidden state vectors directly, leading to a fixed pooling factor, independent of the input sequence length \citep{suganthan2025adaptingdecoderbasedlanguagemodels}.

\begin{wrapfigure}{r}{0.48\textwidth}
    \centering
    \includegraphics[width=0.43\linewidth, trim= 0 0 0 0, clip]{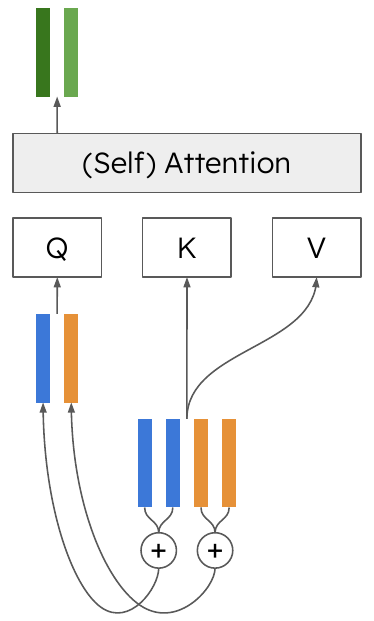}
    \captionof{figure}{\textbf{Pooling in \modelname{$_2$}}. In the last SA, queries are merged by pairs of successive tokens.}
    \label{fig:pooling}
\end{wrapfigure}

We performed an empirical study on how and where to pool tokens to obtain compressed continuous representations. As illustrated in Fig.~\ref{fig:pooling}, pooling is performed in the self-attention module. We average consecutive queries to reach the targeted pooling factor, while keys and values remain unchanged. For a PF of 2 (denoting the encoder as `\modelname{$_2$}'), we group the tokens of the sequence two-by-two. In the self-attention module of the last layer of the encoder LLM, we merge the queries by averaging their continuous hidden states. Then, these pooled queries attend the non-compressed keys and values, mimicking a standard self-attention (SA), but with two-times fewer queries, resulting in a pooling factor of two. We explored inserting the pooling mechanism earlier in the encoder, but this led to poorer performance. This follows the intuition that the information should be as processed as possible before pooling~\citep{tang2025gmsaenhancingcontextcompression}.

\subsection{Training}

\begin{figure*}[t!]
        \centering
        \includegraphics[width=0.85\textwidth]{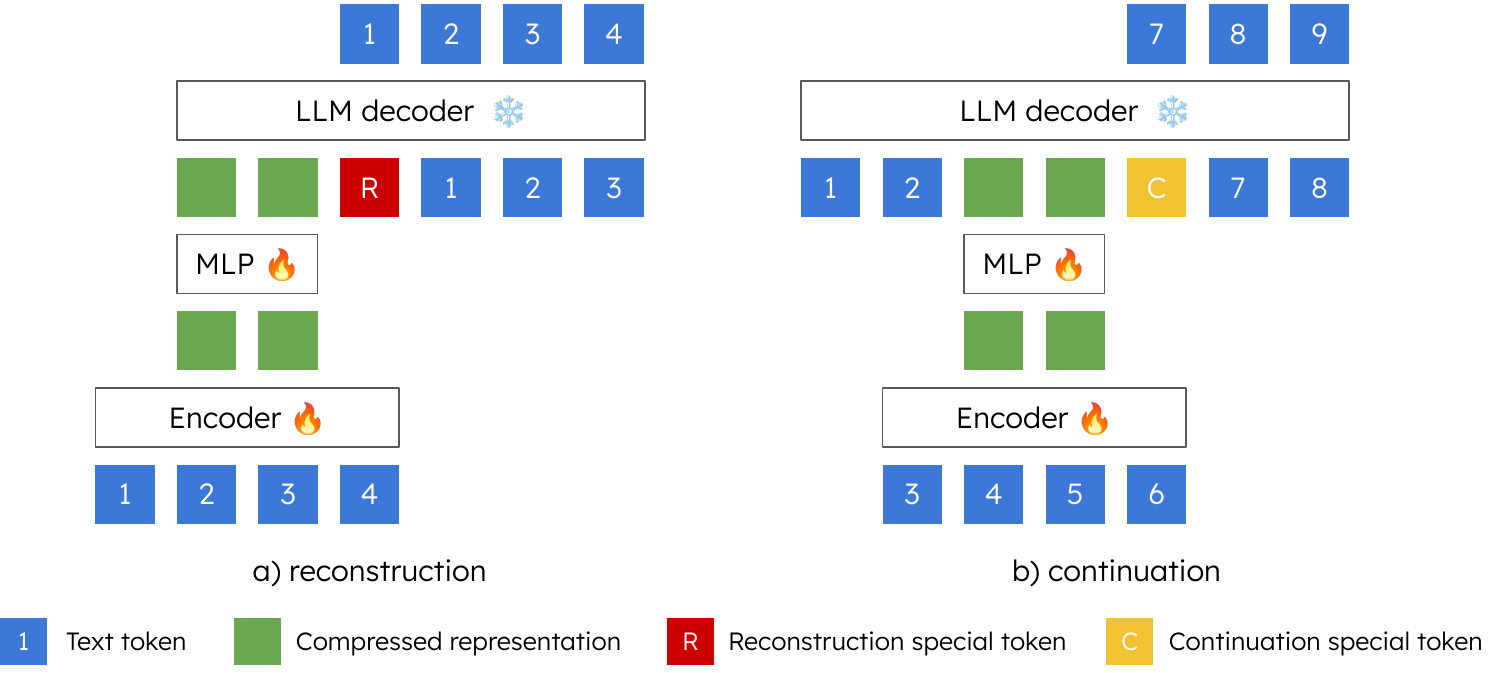}
    \caption{\textbf{\modelname{2} pretraining tasks}. The encoder, special tokens and the MLP are trained through two alternating tasks: a) Reconstruction: compressed tokens are given to the decoder which is teacher-forced to replicate the full text tokens; b) Continuation: a subpart of the tokens in the sequence are compressed and the decoder is teacher-forced to continue starting from the partially compressed sequence.}
    \label{fig:training_tasks}
    \vspace{-1em}
\end{figure*}

\paragraph{Base pretraining.} Memory tokens \citep{ge2024incontextautoencodercontextcompression} show the importance of the reconstruction task when training the model to create easily decompressible continuous representations. However, pure reconstruction is easier than compressing contexts for downstream tasks. In fact, with proper training, our auto-encoder architecture achieves near-perfect reconstruction at a $16 \times$ pooling factor on relatively short sequences (up to around $128$ tokens). Unfortunately, these compressed representations cannot be well exploited by the decoder on downstream tasks, as the model tends to regurgitate the entire context, instead of extracting pertinent information from it. Thus, we consider a second pretraining task, continuation, which is better aligned with inference-time behavior. It consists in replacing subsequences of natural text by their compressed representations, and to teacher-force the continuation immediately following compressed segments. We alternate between these two pretraining tasks as summarized in Fig.~\ref{fig:training_tasks}, and use the standard cross-entropy loss. Additionally, we append special learned tokens \texttt{<Cont>} and \texttt{<Rec>} after each compressed sequence. The two special tokens specify the task (reconstruction or continuation) during training as shown in Fig.~\ref{fig:training_tasks} and enable good downstream results.

\myparag{Fine-Tuning.} After pretraining, our \modelname{} can be flexibly adapted to a range of specific tasks. In particular, it can be used to condition a frozen decoder on task-specific inputs, allowing task specialization without modifying the decoder itself. To preserve the decoder’s few-shot capabilities, we optionally include a small number of in-context examples in the continuation objective--interleaving compressed documents, full-text queries and answers--following a structured prompt template described in Appendix \ref{training_details}. To compute the loss, we mask all tokens except the ones that continue the last compressed sequence, corresponding to the final answer in a few-shot setting. This design encourages generalization and preserves ICL possibilities at inference time with every document being compressed. Furthermore, we demonstrate that zero-shot abilities of instruct decoders can also be preserved; it simply requires to use a different pretraining and fine-tuning template. Once pretrained, \modelname{} can be fine-tuned to master in-context learning, long context understanding, or other downstream tasks. The pooling factor between the two training stages can be changed. It leads to improved models when pretraining with a higher factor than for fine-tuning, see Appendix \ref{sec:comp_rate}. We remind that only the encoder is fine-tuned, and that the decoder is unchanged.

\myparag{Multi-Decoder Training.} To improve the generality of our method, we design a compressor capable of generating token representations that can be used by multiple decoders without any modification. It is a nontrivial challenge due to inherent discrepancies between the hidden-state spaces of different decoder architectures. To overcome this, we employ a shared encoder but we specialize a projector layer (the MLP in Fig.~\ref{fig:training_tasks}) and the special tokens. This set of learned parameters accounts for less than $1\%$ of the encoder’s weights, enabling decoder-specific adaptation with minimal overhead. We find that training with an alternating objective provides the most stable and effective results. At each training step we sample a decoder uniformly and update only its associated projector as well as the shared encoder. This strategy ensures balanced exposure across decoders while maintaining the generalization capacity of the shared encoder.

\section{Experiments}
In this section, we first describe the main settings used for our experiments. Second, we evaluate our method on various downstream tasks that use contexts, such as question answering with retrieved documents or reading comprehension. The contexts for these tasks tend to be short, ranging from 30 to 1,500 tokens. Thus, we also evaluate \modelname{}s for long-context understanding applications. Finally, we report the results of our ablation study on training and architecture as well as an analysis of the memory used to encode Wikipedia with continuous representations from \modelname{}.

\subsection{Experimental setting}
\paragraph{Models.}
We perform experiments using different decoders: Llama3.1 8B \citep{grattafiori2024Llama3herdmodels}, Mistral 7B \citep{jiang2023mistral7b}, both base models, and Llama2 7B Chat \citep{touvron2023Llama2openfoundation}. We design our \modelname{} with Llama3.2 3B as the backbone: we remove its last two layers and no causal mask is applied. The removal of the last two layers is an efficiency-driven design choice. Our ablations show that this truncation decreases encoder size with minimal impact on accuracy. We specifically target the final layers for removal to ensure the compressor is initialized with the most stable hidden states, as opposed to removing earlier layers which would disrupt the feature extraction hierarchy. The MLP consists of two layers: the first projects the 3072-dimensional vectors from the encoder to 2048 dimensions, and the second projects these vectors to 4096-dimensional hidden states, matching the decoder dimension. During inference, we append the special \texttt{<Cont>} token after every sequence of compressed tokens.

\myparag{Training \& Datasets.}
By default, we decide to train all layers of the encoder, including the embedding matrix, using AdamW optimizer \citep{loshchilov2019decoupledweightdecayregularization}. For pretraining, we use data from Common Crawl that has been filtered and processed using dactory\footnote{\url{https://github.com/kyutai-labs/dactory}}. \modelname{} is pretrained on approximately $2.6\text{B}$ tokens. For fine-tuning, we use two different mixes of synthetic and supervised datasets: one for standard context compression benchmarks using base models as decoders and another for long-context benchmarks using an instruct decoder. See the following sections and Appendix \ref{fine_tuning_dataset} for more details.

\subsection{Context compression}

\paragraph{Benchmarks.}
We evaluate our method on question answering, translation, and summarization tasks. For question answering, we use HotpotQA ~\citep[HQA]{yang2018hotpotqadatasetdiverseexplainable} with the \textit{distractor} setting, Natural Questions~\citep[NQ]{47761}, TriviaQA \citep[TQA]{joshi-etal-2017-triviaqa} and SQuAD \citep{rajpurkar2016squad100000questionsmachine}. Top-5 passages are retrieved using NV-Embed v2 \citep{lee2025nvembedimprovedtechniquestraining} from Atlas \citep{izacard2022atlasfewshotlearningretrieval} Wikipedia chunks, simulating a RAG setup. We report exact match (EM) as the main metric, where $\text{EM}=1$ if the normalized predicted and reference answers exactly match. The normalization consists in lowercasing, fixing whitespaces and removing punctuation and articles as standard practices in EM calculations. For translation, we evaluate on FLORES \citep{goyal2021flores101evaluationbenchmarklowresource} averaging BLEU scores across four directions: English to Danish, French, German, and Spanish. Summarization performance is evaluated on CNN-DailyMail using ROUGE-L, which aligns well with human judgments for summarization abilities according to \citet{zhang2023benchmarkinglargelanguagemodels}. All models are evaluated via  5-shot with compressed contexts. Each shot consists of a document, question and answer triplet with the question being either an instruction if the downstream task is a summarization or translation, and a direct question if it is a QA task as illustrated in the templates in Appendix~\ref{eval_data}. The reported pooling factor reflects the per-context ratio of original tokens (using the decoder tokenizer) to compressed tokens. See Appendix \ref{eval_data} for evaluation details.

\myparag{Setting \& Baselines.} Unlike prior works in the context compression literature that often report zero-shot accuracy with instruct decoders—potentially inflating performance when models simply replicate the context—our use of EM aims at better capturing real-world LLM usage by rewarding answers following the few-shot format patterns. We believe it better measures the encoder abilities to produce useful representations, from which the decoder can extract information. Thus, we focus on base model decoders for context compression evaluations. We first set two baselines that reflect the decoder's intrinsic abilities. In the first one, denoted \textit{closed-book}, the decoder relies only on its parametric knowledge and in the second one, called \textit{open-book}, the decoder has access to uncompressed documents in its context.  

For meaningful comparisons, we select a diverse set of strong baselines that capture different approaches to context compression. These include: i) LLMLingua2 \citep{pan2024llmlingua2datadistillationefficient}, which performs hard compression; ii) ICAE \citep{ge2024incontextautoencodercontextcompression}, a soft compression approach using memory tokens; iii) xRAG \citep{cheng2024xragextremecontextcompression}, which relies on pre-computed embeddings for retrieval-augmented generation; iv) PISCO \citep{louis2025piscoprettysimplecompression}, which fine-tunes both the encoder and decoder and states that pretraining is not necessary. We re-implemented the last three baselines using our decoder, fine-tuning dataset, and interleaved fine-tuning task format to ensure a consistent few-shot evaluation setup while preserving each method’s core design. This allows direct comparisons with a common evaluation protocol; See Appendix \ref{baselines_implem} for implementation details.

We fine-tune models on a mix of synthetic translation data and supervised datasets (QA, summarization, paraphrasing, and reading comprehension), explicitly excluding the training sets of our evaluation benchmarks. This setup better highlights our method’s generalization ability (using the train sets of benchmarks strongly improves results, sometimes outperforming the \textit{open-book} results as shown in Tab.~\ref{tab:ft_w_trainsets} of the Appendix). Since each data sample is drawn from one of the sub-datasets, fine-tuning involves stochasticity; unless otherwise specified, we fix the random seed to 0. For our final \modelname{} we follow the best pretraining\,/\,fine-tuning pooling factor pairs as shown in Fig.~\ref{fig:cr_pt_ft_pairs}. This leads to \modelname{$_4$} using the same pretrained encoder as \modelname{$_8$}, only with its fine-tuning performed at a pooling factor of $4$.

\begin{table}[t]
\caption{\textbf{Main comparison of \modelname{} and other models}. `PF' (pooling factor): the token reduction factor (e.g., $4 \times$) for fixed-ratio methods or the number of compressed tokens used, e.g. $\sim 16$, when this number is fixed as this yields benchmark-dependent ratios; `Param.': number of parameters of the encoder;  `Avg. length': mean number of tokens per context document. The superscript on \modelname{} indicates if the model is specifically trained for one decoder ($^M$ for Mistral or $^L$ for Llama) or both simultaneously ($^\otimes$). $^\dagger$ marks modified re-implementations, see details in Appendix \ref{baselines_implem}. Best context compression results are in \textbf{bold}, second best are \underline{underlined}.}
\centering
\small
\setlength{\tabcolsep}{4pt}
\begin{tabular}{c  l l c c c c c c c c}
\toprule
& Methods & PF &Param.& \textbf{NQ} & \textbf{TQA} & \textbf{HQA} & \textbf{SQuAD}& \textbf{FLORES} & \textbf{CNN} & \textbf{Avg.}\\
 \midrule

\multirow{11}{*}{\rotatebox{90}{\textit{Mistral 7B decoder}}} & Avg. length &  &  & $155$ & $152$ & $1479$ & $185$ &   $30$ & $956$ & \\
\tightcmidrule{2-11}
& \textit{closed-book}\cellcolor{gray!10} &$\infty$\cellcolor{gray!10} & \cellcolor{gray!10} & $29.1$\cellcolor{gray!10} &$62.4$\cellcolor{gray!10} &$22.8$\cellcolor{gray!10} &$17.1$\cellcolor{gray!10} & \cellcolor{gray!10} & \cellcolor{gray!10} & \cellcolor{gray!10}\\
&\textit{open-book}\cellcolor{gray!10} &$1 \times$\cellcolor{gray!10} & \cellcolor{gray!10} &$39.9$\cellcolor{gray!10} & $70.5$\cellcolor{gray!10} & $48.3$\cellcolor{gray!10} &$77.7$\cellcolor{gray!10} & $31.3$\cellcolor{gray!10} & $27.2$\cellcolor{gray!10} & $49.2$\cellcolor{gray!10}\\
& ICAE-like$^\dagger$  &  $\sim 32$ & $7.2$B  & $36.5$ & $66.7$ & $24.3$ & $58.8$ & $28.3$ &$15.8$ & $38.4$\\
& xRAG-like$^\dagger$  &  $\sim 1$ & $7.1$B & $30.7$ & $65.2$ & $21.5$ & $23.9$ & $0.9$ &$14.6$&$26.1$\\
& LLMLingua2 &  $1.9 \times$  & $0.6$B & $\underline{38.8}$ & $\underline{69.0}$ & $\underline{43.7}$ & $59.2$ &   $12.6$& $\underline{24.9}$& $41.4$\\
& PISCO-like$^\dagger$& $\sim 32$ & $7.2$B  & $34.7$ & $68.5$ & $24.9$ & $38.2$ & $\underline{33.6}$ & $19.2$& $36.5$\\
& &$4 \times$ & $-$  & $36.6$ & $69.2$ & $29.4$ & $48.1$ & $\mathbf{34.5}$ & $19.3$& $39.5$\\
\tightcmidrule{2-11}
& \modelname{$_4$}$^\otimes$\cellcolor{blue!8}    &$4 \times$\cellcolor{blue!8}    & $3.0$B  \cellcolor{blue!8} & $38.2$\cellcolor{blue!8}    &$\mathbf{70.4}$\cellcolor{blue!8}    &$40.8$\cellcolor{blue!8}    &$\underline{69.2}$\cellcolor{blue!8}    & $29.5$\cellcolor{blue!8}    &$\mathbf{25.6}$\cellcolor{blue!8}    & $\underline{45.6}$\cellcolor{blue!8}   \\
&\modelname{$_4$}$^M$\cellcolor{blue!8}    & $4 \times$\cellcolor{blue!8}    & $-$\cellcolor{blue!8}    & $\mathbf{39.0}$\cellcolor{blue!8}    &$68.9$\cellcolor{blue!8}    & $\mathbf{45.1}$\cellcolor{blue!8}    &  $\mathbf{71.1}$\cellcolor{blue!8}    & $31.0$\cellcolor{blue!8}    & $23.8$\cellcolor{blue!8}    & $\mathbf{46.5}$\cellcolor{blue!8}   \\
&\modelname{$_8$}$^M$\cellcolor{blue!8}    &$8 \times$\cellcolor{blue!8}    &$-$\cellcolor{blue!8}    & $38.4$\cellcolor{blue!8}    & $67.9$\cellcolor{blue!8}    & $40.8$\cellcolor{blue!8}    & $62.0$\cellcolor{blue!8}    & $28.3$\cellcolor{blue!8}    & $22.9$\cellcolor{blue!8}    & $43.4$\cellcolor{blue!8}   \\
\midrule
\multirow{11}{*}{\rotatebox{90}{\textit{Llama3.1 8B decoder}}} & Avg. length &  &  & $135$ & $133$ & $1285$ & $164$ & $27$  & $855$ &\\
\tightcmidrule{2-11}
&\textit{closed-book}\cellcolor{gray!10} &$\infty$\cellcolor{gray!10} & \cellcolor{gray!10} & $25.4$\cellcolor{gray!10} & $60.6$\cellcolor{gray!10} & $21.6$\cellcolor{gray!10} & $15.3$\cellcolor{gray!10} &    \cellcolor{gray!10} & \cellcolor{gray!10} & \cellcolor{gray!10} \\
&\textit{open-book}  \cellcolor{gray!10} &  $1 \times$\cellcolor{gray!10} & \cellcolor{gray!10} &  $38.6$\cellcolor{gray!10} &  $67.1$\cellcolor{gray!10} &  $47.1$\cellcolor{gray!10} & $72.2$\cellcolor{gray!10} & $32.8$\cellcolor{gray!10} & $26.5$\cellcolor{gray!10} & $47.4$\cellcolor{gray!10} \\
& ICAE-like$^\dagger$  &  $\sim 32$& $7.6$B & $38.4$ &  $67.3$ & $20.5$ & $61.6$ & $31.3$ & $17.3$& $39.4$\\
& xRAG-like$^\dagger$ &  $\sim 1$& $7.1$B & $28.0$ & $62.1$ & $21.7$ & $22.3$ &  $3.4$&$12.7$&$25.0$\\
& LLMLingua2 &  $2.0 \times$ & $0.6$B & $36.1$ & $66.3$ & $\underline{45.2}$ & $58.8$ &  $13.6$& $\underline{23.8}$ & $40.6$\\
& PISCO-like$^\dagger$ & $\sim 32$ & $7.6$B & $35.1$ &  $69.4$ & $30.6$  & $40.5$ & $\underline{35.2}$ & $19.7$ & $38.4$\\
& & $4 \times$& $-$ &  $37.9$ & $\underline{70.5}$ & $37.0$ & $57.2$ & $\mathbf{36.5}$& $20.7$ & $43.3$\\
 \tightcmidrule{2-11}
& \modelname{$_4$}$^\otimes$\cellcolor{blue!8}    & $4 \times$\cellcolor{blue!8}    & $3.0$B\cellcolor{blue!8} & $\underline{39.6}$\cellcolor{blue!8}    & $\mathbf{70.8}$\cellcolor{blue!8}    & $43.6$\cellcolor{blue!8}    & $\underline{71.8}$\cellcolor{blue!8}    & $32.8$\cellcolor{blue!8}    & $\mathbf{26.1}$\cellcolor{blue!8}    & $\underline{47.5}$\cellcolor{blue!8}   \\
& \modelname{$_4$}$^L$\cellcolor{blue!8}    &  $4 \times$\cellcolor{blue!8}    & $-$\cellcolor{blue!8}    & $\mathbf{39.7}$\cellcolor{blue!8}    & $70.1$\cellcolor{blue!8}    & $\mathbf{46.9}$\cellcolor{blue!8}    & $\mathbf{74.0}$\cellcolor{blue!8}    & $33.7$\cellcolor{blue!8}    & $23.7$\cellcolor{blue!8}    & $\mathbf{48.0}$\cellcolor{blue!8}   \\
& \modelname{$_8$}$^L$\cellcolor{blue!8}    &  $8 \times$\cellcolor{blue!8}    & $-$\cellcolor{blue!8}    & $38.9$\cellcolor{blue!8}    & $69.0$\cellcolor{blue!8}    & $42.8$\cellcolor{blue!8}    & $66.0$\cellcolor{blue!8}    & $30.6$\cellcolor{blue!8}    & $22.8$\cellcolor{blue!8}    &  $45.0$\cellcolor{blue!8}   \\
\bottomrule
\end{tabular}
\label{table:main_results}
\end{table}

\myparag{Results.} In Tab.~\ref{table:main_results}, we report our main results on context compression using the best checkpoints of \modelname{}. Specifically, we use the model pre-trained on 2.6B tokens, as opposed to the 2B-token variant used in our ablation studies. First, we observe that our models outperform the \textit{closed-book} baseline, showing that frozen decoders can extract useful information for downstream tasks from the compressed tokens. We note that \modelname{} tends to outperform the baselines more on tasks where the decoder cannot rely on its own parametric memory, such as reading comprehension (SQuAD) or summarization (CNN). However, \modelname{} is slightly behind PISCO-like methods on FLORES because those sequences are very short ($<30$ tokens). For methods using 32 fixed memory tokens, this results in a compression ratio below 1.0 even with chunking strategies, effectively expanding the input rather than compressing it and providing more information capacity than the open-book baseline. Conversely, \modelname{} maintains a constant pooling factor, ensuring true compression efficiency regardless of sequence length. Furthermore, we observe that both specialized and shared \modelname{} perform better when paired with the Llama3.1 8B decoder, likely because it belongs to the same model family as our encoder backbone based on Llama3.2 3B. Finally, \modelname{$_4$} nearly matches the \textit{open-book} baseline without altering the decoder and while achieving $\times 1.8$ gains of prefilling FLOPs, as profiled in \textsection~\ref{subsec:profiling_exps}, larger pooling factors in Appendix~\ref{subsub:icae_further_comparisons}. Please note that for certain baselines, our results differ from the ones reported in previous work. This is due to the different setting that we consider in this paper: 1) using exact match as the metric, 2) excluding the training sets of the benchmarks from the fine-tuning data and 3) using base models instead of instruct ones.

\myparag{Encoder adaptation to multi-decoder.}
Through multiple experiments, we found that a single encoder
can be trained to be used by multiple decoders. While we experiment with two decoders for engineering simplicity, our adaptation to new decoders experiments suggest that the method can be extended to more than two decoders. More specifically, during both pretraining and fine-tuning, we use a joint learning setup where at each training step, we sample which decoder to use among the two targets. To improve performance, we introduce separate MLP projectors and special task tokens for each decoder. This allows lightweight specialization of compressed representations, adding only $15$M parameters per decoder. We report results in Tab.~\ref{table:main_results}, showing that on average, the common encoder, \modelname{}$^\otimes$, loses less than $1.0$ point compared to its specialized counterparts. More encoder-decoder pairings are tested in Appendix \ref{sec:encoder_pairing}.\\

\textbf{Adaptation to new decoders.} Once our \modelname{} has been trained for a decoder, we can adapt it to new decoders with minimal adjustments. Indeed, we solely learn a new MLP projector and special tokens to feed a third decoder, OLMo-7B \citep{Groeneveld2023OLMo}, while keeping the encoder frozen. Fine-tuning leads to better results than the \textit{closed-book} baseline by only training $15$M parameters. However, the score gap with the \textit{open-book} setting remains relatively larger than when training with the alternating decoder objective as shown in Tab.~\ref{table:main_results} with Llama and Mistral models. Interestingly, on benchmarks where the decoder was limited by its context window, such as HotpotQA, using \modelname{$_4$} outperforms the \textit{open-book} setup.

\begin{table}[htbp]
\caption{
\textbf{Adaptation to a new decoder}. 
Due to OLMo 7B $2048$-token context window, we truncate documents for \textit{open-book} baseline to $400$ tokens. Per decoder specific parameters are reported.}
\centering
\small
\setlength{\tabcolsep}{4pt}
\begin{tabular}{c l l c c c c c c c c}
\toprule
 & Methods & PF & Param.& \textbf{NQ} & \textbf{TQA} & \textbf{HQA} & \textbf{SQuAD} &  \textbf{FLORES} & \textbf{CNN} &\textbf{Avg.}\\
\midrule
\multirow{4}{*}{\rotatebox{90}{\small \textit{OLMo-7B}}} &  \textit{closed-book} \cellcolor{gray!10} &$\infty$\cellcolor{gray!10} &\cellcolor{gray!10} & $19.5$\cellcolor{gray!10} & $48.3$\cellcolor{gray!10} & $17.8$\cellcolor{gray!10} & $11.2$\cellcolor{gray!10} &  \cellcolor{gray!10} &\cellcolor{gray!10} & \cellcolor{gray!10} \\
& \textit{open-book} \cellcolor{gray!10} &  $1\times$\cellcolor{gray!10} & \cellcolor{gray!10} &  $35.3$\cellcolor{gray!10} & $64.8$\cellcolor{gray!10} & $22.9$\cellcolor{gray!10} & $67.9$\cellcolor{gray!10} &   $22.2$\cellcolor{gray!10} & $24.2$\cellcolor{gray!10} & $39.6$\cellcolor{gray!10}\\
&\modelname{$_4$}$^M$\cellcolor{blue!8}    &$4\times$\cellcolor{blue!8}    &$15$M\cellcolor{blue!8} & $31.5$\cellcolor{blue!8}    & $62.5$\cellcolor{blue!8}    & $26.5$\cellcolor{blue!8}    &  $46.4$\cellcolor{blue!8}    & $17.2$\cellcolor{blue!8}    & $19.1$\cellcolor{blue!8}    & $33.9$\cellcolor{blue!8}   \\
&\modelname{$_4$}$^\otimes$ \cellcolor{blue!8} & $4\times$\cellcolor{blue!8}    & $-$\cellcolor{blue!8}    & $33.1$\cellcolor{blue!8}    & $63.1$\cellcolor{blue!8}    & $25.0$\cellcolor{blue!8}    & $44.6$\cellcolor{blue!8}    & $17.1$\cellcolor{blue!8}    &  $18.9$\cellcolor{blue!8}    & $33.6$\cellcolor{blue!8}    \\
\bottomrule
\end{tabular}
\label{table:result_add_decoder}
\end{table}

\subsection{Long Context}
In this section, we adapt our fine-tuning method to handle longer contexts, testing our architecture on long-context understanding tasks. We pretrain and fine-tune an \modelname{$_8$} paired with an instruct decoder Llama2 7B Chat \citep{touvron2023Llama2openfoundation}. During fine-tuning, we encode fixed-size chunks of long documents in parallel and feed them to the decoder by concatenating the compressed tokens of each chunk. For fine-tuning, we synthesize a QA and summarization dataset based on concatenated Wikipedia chunks, PG-19 books \citep{raecompressive2019} and ArXiv papers from RedPajama \citep{weber2024redpajama} using Gemma3-27B \citep{gemmateam2025gemma3technicalreport}. Then, we divide each context in up to $32$ chunks of $1024$ tokens. Similarly at inference, contexts are truncated to $32$k tokens and then \modelname{$_8$} processes chunks in parallel. In this setting, we remove the special tokens, as instruction prompts play a similar role, see Appendix~\ref{training_details} for technical details.

\myparag{Benchmarks \& Models.}
For long-context understanding, we report F1 score on NarrativeQA and QASPER and report Rouge-L on GovReport and QM-Sum validation sets from the ZeroSCROLLS benchmark~\citep{shaham2023zeroscrollszeroshotbenchmarklong}, a suite of zero-shot long-context understanding tasks. Specifically, we adopt the task formats and instructions as used in \citet{yen2024longcontextlanguagemodelingparallel}\footnote{\url{https://github.com/princeton-nlp/CEPE}}. On these benchmarks, we compare our model to Llama2 7B Chat, which is constrained by a context window of $4096$ tokens, as well as to open-source models specifically designed to extend its limited context window. These include Llama2-32k Instruct \citep{together2023Llama2_7b_32k_instruct}, which relies on positional interpolation combined with fine-tuning, and CEPED \citep{yen2024longcontextlanguagemodelingparallel}, which employs a lightweight encoder to process chunks of input in parallel, feeding their representations into a decoder through learned cross-attention layers.

\begin{table}[htbp]
\caption{\textbf{Long-context evaluation on long-context benchmarks}. The token count includes all tokens fed to the decoder. 
CEPED uses 2k decoder tokens plus encoder-side tokens.}
\centering
\setlength{\tabcolsep}{4pt}
\begin{tabular}{l c c c c c}
\toprule
Models & Max. Tokens & \textbf{NQA} & \textbf{Qspr} & \textbf{GvRp} & \textbf{QM-Sum} \\
\midrule
Llama2 Chat & $4$k  & $16.1$&$17.2$ & $15.7$ & $19.8$ \\
\midrule
+ CEPED & $2\text{k}+30\text{k}$ &$20.5$ & $19.7$& $12.7$ & $19.7$\\
Llama2-32k Instruct & $32$k  &  $14.2$& $16.4$& $17.8$& $17.6$\\
\rowcolor{blue!8}
\modelname{$_8$} + Llama2 Chat & $4$k $(32\text{k}//8)$& $27.5$& $28.3$& $14.1$& $19.1$\\
\bottomrule
\end{tabular}
\label{tab:long_ctx}
\end{table}

\myparag{Results.} In Tab.~\ref{tab:long_ctx}, we show that feeding compressed tokens from \modelname{$_8$} to Llama2 Chat improves long-context QA performance. It allows the model to process up to $8\times$ more input than its original context window. This also shows that the decoder can interpret the compressed tokens without any parameter modification. Thus, a small model’s context window can be extended simply by training an external compressing encoder. On some tasks, our approach even outperforms methods that expand Llama2’s context window  through new learned internal modules or full-model fine-tuning. This advantage likely stems from our synthetic fine-tuning dataset, which reflects the answer-length distributions of our benchmarks. Conversely, our underperformance on GvRp and QM-Sum suggests a distribution mismatch, as these tasks differ from our training data. This highlights \modelname{}'s capacity for task-specific context extension while underscoring the necessity of well-aligned fine-tuning data. In Appendix \ref{subsec:niah}, we show  that ARC-Encoder achieves strong and more consistent fine-grained retrieval performance across varying context lengths than other tested models. Crucially, our approach is unique in leaving the decoder unchanged, ensuring consistent behavior across standard tasks while significantly enhancing long-context understanding. While we demonstrate this capability by extending Llama2, we leave scaling to longer contexts (e.g., 128k tokens) for future work, as such an expansion would necessitate specialized datasets and substantial computational budgets.

\subsection{Compression Ablations}
\label{sec:ablation}

In this section, we discuss key design choices of our method. We compare context compression results using the same evaluation setting as in Tab.~\ref{table:main_results}, mostly showing the average score on all these benchmarks. Unless stated otherwise, models are pretrained and then fine-tuned using a pooling factor of $8$ with the Mistral 7B decoder. To reduce the computation costs of these ablations we pretrained on approximately $2$B tokens only, roughly $75\%$ of the tokens used for models in Tab.~\ref{table:main_results}. Additional details are provided in Appendix \ref{ablation_default}.

\begin{minipage}[t]{0.71\textwidth}
\paragraph{Training objective.} In contrast to \citet{louis2025piscoprettysimplecompression}, we show that pretraining is essential for our approach. While fine-tuning is also crucial, it cannot provide competitive results on downstream tasks on its own. Long pretraining is essential for aligning \modelname{} outputs with the decoder’s hidden state space. For example, after $20$k pretraining steps, we observe an improvement of approximately $+16$ points on the average score, while after $80$k steps, the improvement reaches $+19$ points, compared to directly fine-tuning without pretraining. Without fine-tuning, the decoder fails to use the compressed context leading to large performance drop in translations, reading comprehension and summarization.

\end{minipage}%
\hfill
\begin{minipage}[t]{0.26\textwidth}
\vspace{-2em}
\begin{table}[H]
\centering
\caption{Impact of pretraining reconstruction ratio in $\%$ of the sampled batches.}
\setlength{\tabcolsep}{4pt}
\begin{tabular}{l c}
\toprule
\textbf{$\%$ Rec.} & \textbf{Avg.}\\
\midrule
$0\%$ & $39.8$ \\
$20\%$ & $\textbf{41.6}$ \\
$50\%$ & $41.5$ \\
$100\%$ & $37.5$ \\
\bottomrule
\end{tabular}
\label{table:perc_rec}
\end{table}
\end{minipage}\\

Beyond pretraining length, the choice of pretraining tasks also proves critical: Tab.~\ref{table:perc_rec} shows that omitting reconstruction or using too little continuation leads to substantial performance drops. In practice, early training updates mainly reduce the reconstruction loss, suggesting that the encoder and MLP first learn to produce correctly aligned compressed representations with the embedding space of the decoder, after which the continuation objective encourages representations that the LLM can use effectively for downstream generation. We also tried adding context distillation as in \citet{cheng2024xragextremecontextcompression} during fine-tuning but it did not improve results while adding a large computational overhead. 

\begin{table}[t]
\caption{\textbf{Ablations on encoder design}. \emph{Default setting} corresponds to \modelname{$_8$}$^M$ with only $60$k pretraining steps. Results are averaged over 3 seeds, with standard deviations shown in indices. }
\centering
\setlength{\tabcolsep}{4pt}
\begin{tabular}{l c c c c c c c c}
\toprule
 & Param.& \textbf{NQ} & \textbf{TQA} & \textbf{HQA} & \textbf{SQuAD}  & \textbf{FLORES} & \textbf{CNN} & \textbf{Avg.}\\
\midrule
\textbf{\textit{Default setting}}  &  $3.0$B   &  $36.9_{\pm0.2}$ & $67.2_{\pm0.3}$ & $39.9_{\pm0.4}$ & $58.3_{\pm2.0}$ & $27.4_{\pm0.3}$ & $20.1_{\pm0.1}$ & $41.7_{\pm0.4}$\\
\midrule
\multicolumn{8}{c}{\textit{How to truncate the encoder?}} \\
\midrule
Truncate 0 layer &  $3.2$B & $38.8_{\pm0.3}$ & $67.7_{\pm0.3}$ & $39.4_{\pm0.9}$ & $61.6_{\pm0.8}$ & $26.9_{\pm0.2}$ & $20.3_{\pm0.1}$ & $42.4_{\pm0.3}$\\
Truncate 4 layers &  $2.8$B & $37.6_{\pm0.3}$ & $67.5_{\pm0.3}$ & $39.1_{\pm0.6}$ & $60.1_{\pm1.3}$ & $25.0_{\pm0.6}$ & $20.4_{\pm0.3}$ & $41.6_{\pm0.3}$ \\
Truncate 21 layers &  $1.1$B & $37.0_{\pm0.3}$ & $67.3_{\pm0.1}$ & $32.5_{\pm0.2}$ & $52.1_{\pm0.5}$ & $23.5_{\pm0.2}$ & $19.2_{\pm0.3}$&  $38.6_{\pm0.1}$\\
\midrule
\multicolumn{8}{c}{\textit{How to pool?}} \\
\midrule
by 2 every last layers &  $3.0$B & $38.2_{\pm0.6}$ & $68.4_{\pm0.4}$ & $40.0_{\pm0.2}$ & $61.1_{\pm0.6}$ & $26.7_{\pm0.3}$ & $20.1_{\pm0.1}$ &$42.4_{\pm0.2}$\\
by 2 every two layers &  $-$ & $38.3_{\pm0.2}$ & $68.1_{\pm0.6}$ & $38.9_{\pm0.4}$ & $61.1_{\pm1.3}$ & $26.4_{\pm0.2}$ & $20.0_{\pm0.2}$& $42.1_{\pm0.3}$\\
\midrule
\multicolumn{8}{c}{\textit{How to modify the encoder?}} \\
\midrule
w LoRA (all, rank$=128$) &  $-$ & $38.8_{\pm0.4}$ & $67.0_{\pm0.5}$ & $37.0_{\pm0.2}$ & $57.7_{\pm1.4}$ & $27.0_{\pm0.1}$ & $19.1_{\pm0.5}$ & $41.1_{\pm0.4}$\\
w causality &  $-$ & $38.7_{\pm0.3}$ & $67.9_{\pm0.6}$ & $37.4_{\pm0.3}$ & $57.3_{\pm1.1}$ & $27.0_{\pm0.3}$ & $19.4_{\pm0.1}$ & $41.3_{\pm0.2}$\\
\bottomrule
\end{tabular}
\label{tab:encoder_design}
\end{table}

\myparag{Encoder Architecture.}
This ablation explores design choices for reducing encoder size or boosting performance. Thanks to strong results in multi-decoder and long-context applications, we keep the default setting architecture in Tab.~\ref{table:main_results} for consistency reasons, though other designs may excel in other use cases. Tab.~\ref{tab:encoder_design} shows the trade-off between encoder parameters and performance when truncating layers of the LLM backbone. We observe the same trade-off when using LoRA instead of full fine-tuning. We maintained full fine-tuning as our default setting to minimize the number of hyperparameters. Alternative pooling schedules also look promising: pooling every two tokens in the last layers performs better than our default at the same pooling factor, suggesting pooling strategies could be adapted to the target pooling factor. Adding special learned tokens is particularly useful to help the encoder generalize to new decoders. We chose to remove causality from our standard configuration because it limits the exploration of non-causal pooling strategies and results in slightly worse performance on reading comprehension tasks without providing significant benefits.

\begin{table}[H]
\centering
\caption{\textbf{Different pooling methods}. Scores are averaged over $3$ fine-tunings with different seeds.}
\setlength{\tabcolsep}{4pt}
\begin{tabular}{c l c  c c c c c c c}
\toprule
& Pooling method & PF & \textbf{NQ} & \textbf{TQA} & \textbf{HQA} & \textbf{SQuAD} & \textbf{FLORES} & \textbf{CNN} & \textbf{Avg.}\\
\midrule
\multirow{6}{*}{\rotatebox{90}{\textit{Mistral 7B}}}&Average queries & $\times 8$&  $36.9_{\pm0.2}$ & $67.2_{\pm0.3}$ & $39.9_{\pm0.4}$ & $58.3_{\pm2.0}$ & $27.4_{\pm0.3}$ & $20.1_{\pm0.1}$ & $41.7_{\pm0.4}$\\
& $-$ & $\times 4$&  $38.7_{\pm0.1}$ & $67.5_{\pm0.6}$ & $42.0_{\pm0.6}$ & $71.9_{\pm0.4}$& $29.3_{\pm0.2}$ & $23.5_{\pm1.3}$ & $45.5_{\pm0.1}$\\
&Last queries & $\times 8$ & $37.6_{\pm0.3}$ & $68.1_{\pm0.3}$ & $39.7_{\pm0.4}$ & $59.8_{\pm1.0}$ & $27.3_{\pm0.2}$ & $19.7_{\pm0.2}$ & $42.1_{\pm0.3}$\\
&$-$ & $\times 4$ & $38.2_{\pm0.5}$ & $67.1_{\pm0.6}$ & $40.9_{\pm0.3}$ & $71.1_{\pm0.4}$ & $28.8_{\pm0.1}$ & $21.1_{\pm0.5}$ & $44.5_{\pm0.2}$\\
&Kmeans merged queries & $\times 8$ &  $37.1_{\pm0.6}$ & $65.8_{\pm0.2}$ & $38.4_{\pm0.2}$ & $52.3_{\pm0.2}$ & $18.6_{\pm0.3}$ & $19.4_{\pm0.6}$ & $38.6_{\pm0.1}$\\
&Memory tokens & $\sim 16$ &  $36.8_{\pm0.6}$ & $66.8_{\pm0.9}$ & $28.9_{\pm0.4}$ & $49.8_{\pm1.2}$ & $31.0_{\pm0.3}$ & $17.0_{\pm0.2}$ & $38.4_{\pm0.1}$\\
\bottomrule
\end{tabular}
\label{tab:pooling_methods_ablations}
\end{table}

\myparag{How to pool?}The pooling operation should merge information from continuous representations while still producing vectors interpretable by the decoder. We experiment with memory tokens, the standard approach in the field. This method performs poorly as sequence length increases (see Tab.~\ref{tab:pooling_methods_ablations}), since the effective compression becomes higher. We also tested k-means clustering, but this merging of potentially distant tokens proved harmful 
especially in translation tasks. It assumes in addition a non-causal training of the encoder, failing completely without it. The most effective poolings instead use contiguous tokens, either by averaging them as in Fig.~\ref{fig:pooling}, or by selecting the last token of each segment. Since averaging proved more robust in the multi-decoder setting, we choose it as our default method. While we explored dynamic merging strategies to achieve a specific average pooling factor, these initial attempts yielded variable results, suggesting that further refinement of the merging logic is required.

\begin{table}[H]
\centering
\caption{\textbf{Ablation on pooling module placement.} Scores are averaged over three fine-tuning runs with different seeds. All methods rely on averaging continuous representations.}
\setlength{\tabcolsep}{4pt}
\begin{tabular}{c l c  c c c c c c c}
\toprule
& Pooling method & PF & \textbf{NQ} & \textbf{TQA} & \textbf{HQA} & \textbf{SQuAD} & \textbf{FLORES} & \textbf{CNN} & \textbf{Avg.}\\
\midrule
\multirow{4}{*}{\rotatebox{90}{\textit{Mistral 7B}}}&\textbf{\textit{Default setting}} & $8\times$&  $36.9_{\pm0.2}$ & $67.2_{\pm0.3}$ & $39.9_{\pm0.4}$ & $58.3_{\pm2.0}$ & $27.4_{\pm0.3}$ & $20.1_{\pm0.1}$ & $41.7_{\pm0.4}$\\
& Between SA and MLP & $-$ & $36.6_{\pm1.1}$ & $63.4_{\pm0.5}$ & $39.1_{\pm0.6}$ & $58.5_{\pm0.9}$ & $26.0_{\pm0.2}$ & $21.0_{\pm0.6}$ & $40.8_{\pm0.3}$\\
& Before queries & $-$ &  $37.0_{\pm0.6}$ & $64.6_{\pm0.1}$ & $37.8_{\pm0.4}$ & $60.2_{\pm0.1}$ & $26.3_{\pm0.2}$ & $22.4_{\pm0.4}$ & $41.4_{\pm0.2}$\\
& 2 layers earlier & $-$ &  $37.0_{\pm0.1}$ & $65.0_{\pm0.4}$ & $39.4_{\pm0.8}$ & $60.6_{\pm0.9}$ & $26.5_{\pm0.2}$ & $21.7_{\pm0.3}$ & $41.7_{\pm0.4}$\\
\bottomrule
\end{tabular}
\label{tab:pooling_methods_place_ablations}
\end{table}

\myparag{Where to pool?} Pooling module placement is a critical architectural choice. To isolate its impact, we evaluate Mistral 7B at an $\times 8$ pooling factor, consistently applying average pooling adjusted to each placement. We compare our default setting (\textsection~\ref{subsec:pooling_method}) against three alternatives: (1) \textit{Between SA and MLP}, averaging hidden states strictly between the final layer's self-attention and feed-forward blocks; (2) \textit{Before queries}, merging prior to the final layer's self-attention; and (3) \textit{2 layers earlier}, shifting the default pooling two layers deeper. As Tab.~\ref{tab:pooling_methods_place_ablations} shows, altering intra-layer placement slightly degrades average performance. Interestingly, pooling two layers earlier yields an identical average score, indicating robustness to slightly earlier pooling. However, shifting the module further upstream to the middle of the encoder drops performance by over $1\%$ on average. Furthermore, at a $\times 4$ pooling factor, the default setting outperforms the \textit{2 layers earlier} variant by $0.5\%$, justifying our final choice.

\subsection{Profiling}
\label{subsec:profiling_exps}
\begin{figure*}[!ht]
    \centering
    \begin{subfigure}[t]{0.35\textwidth}
        \centering
        \includegraphics[width=\linewidth]{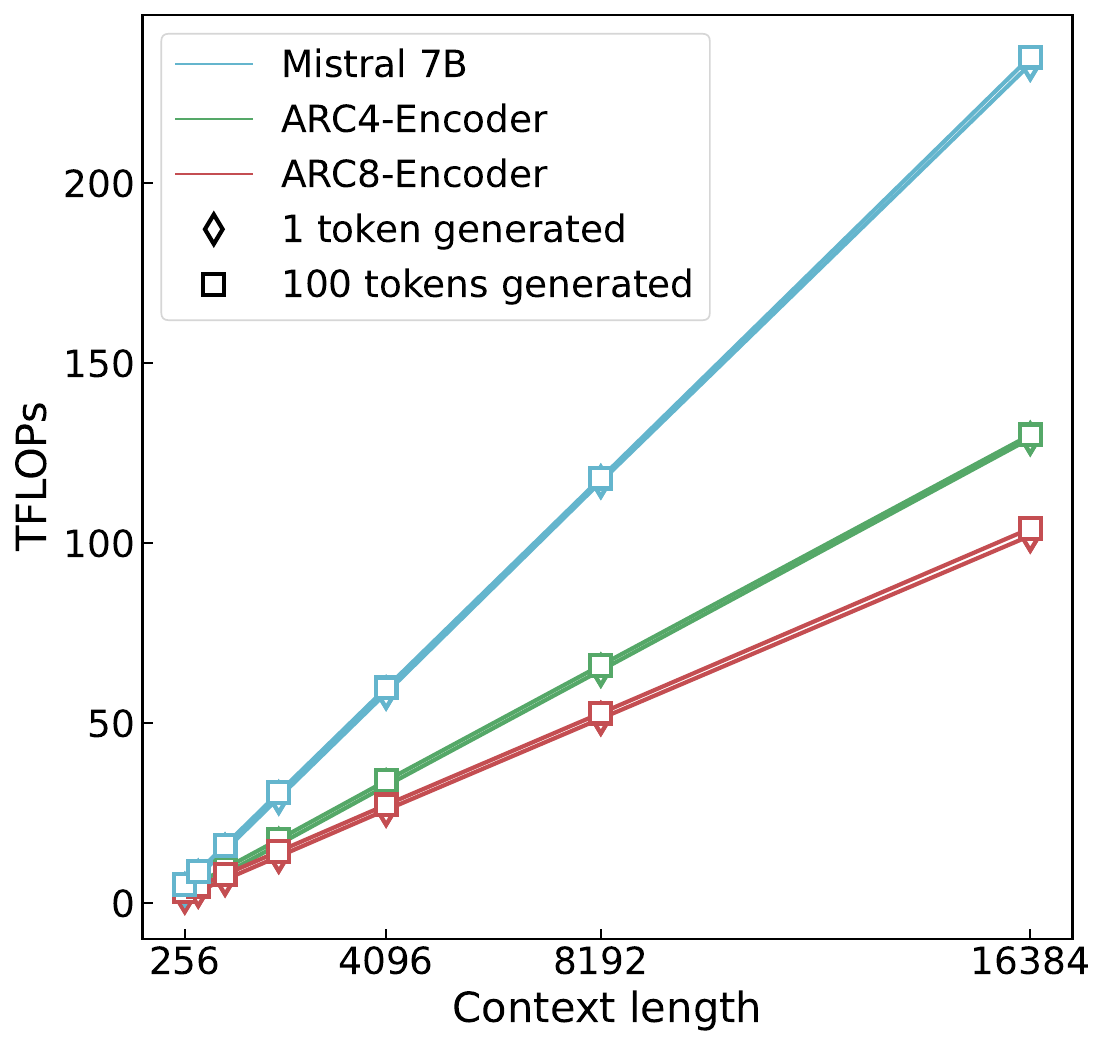}
        \caption{Number of TFLOPs}
    \end{subfigure}%
    \hspace{0.1\textwidth}
    \begin{subfigure}[t]{0.35\textwidth}
        \centering
        \includegraphics[width=\linewidth]{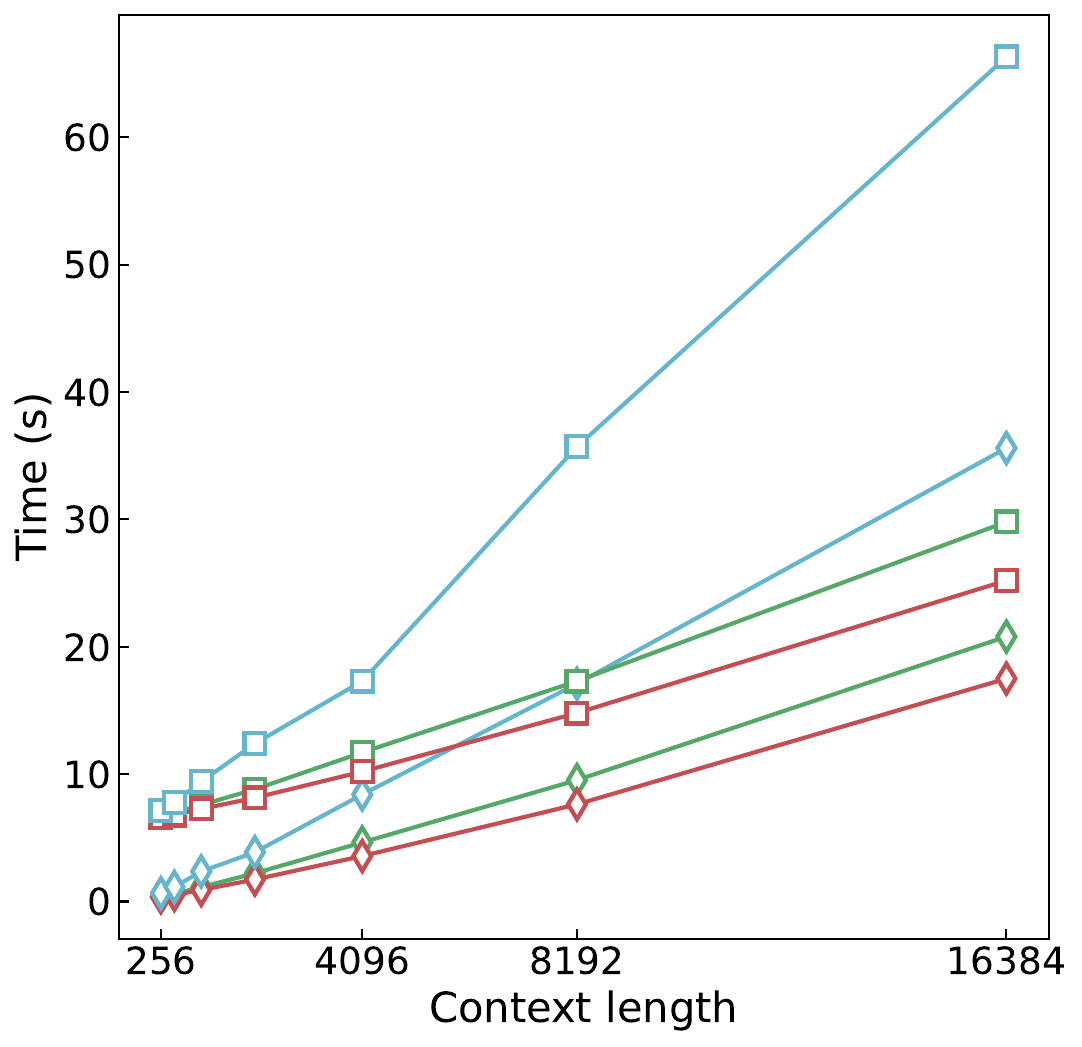}
        \caption{CUDA time (s)}
    \end{subfigure}
    \caption{\textbf{Measured computational costs}. (a) TFLOPs and (b) CUDA time in seconds for the continuation of a book (PG19) for various prompt lengths and numbers of tokens to generate on one NVIDIA H100.}
    \label{fig:flops_time_analysis}
\end{figure*}

By compressing context we aim at speeding up inference as well as reducing the computations. In practice, we evaluate these benefits using Torch Profiler\footnote{\url{https://docs.pytorch.org/tutorials/recipes/recipes/profiler_recipe.html}} measuring CUDA execution time (s) and tera FLOPs (TFLOPS) for the compression, prefilling and decoding stages with various prompt context lengths and various numbers of tokens to generate. All experiments use Mistral 7B decoder and are run in float32 on one NVIDIA H100 GPU with a batch size of 1. We force the decoder to continue its context prompt, compressed or not, and generate $n \in \{1,100\}$ tokens. As shown in Fig.~\ref{fig:flops_time_analysis}, generation is less costly in terms of compute when using compressed tokens by \modelname{}. The compute cost of compression is amortized during the prefilling phase since the decoder has fewer tokens to process. In terms of memory, the compressed representations could be pre-computed on another GPU as shown in \textsection~\ref{subsection:memory_analysis}. It would use $6$GB of memory in bfloat16 for our standard model but could be reduced by training smaller \modelname{} as experimented in Appendix~\ref{subsection:smaller_encoder}.
\subsection{Memory Analysis}
\label{subsection:memory_analysis}
\begin{minipage}{0.59\textwidth}
    When compressing contexts on the fly, \modelname{$_4$} already leads to a $1.8\times$ speed-up compared to using the natural text. In the case where contexts are potentially used multiple times, such as in RAG systems, even greater speed-ups could be achieved by \emph{pre-computing} the compressed representations and storing them. This option is only viable if the size of compressed representations is roughly the same as the original text. Here, we explore the following tradeoffs to reduce the size of compressed representations: 1) changing the pooling factor, 2) reducing the dimension of the MLP bottleneck and 3) quantizing the representations using product quantization~\citep[PQ]{jegou:inria-00514462}, by increasing the dimension of the sub-quantizers while keeping the number of centroids fixed. We report results in Fig.~\ref{fig:memory_bottleneck}, showing that by combining these different methods (bottleneck dimension varies within curves of the same color, and markers indicate the pooling factor), encoding English Wikipedia with \modelname{} requires 80 GiB with minimal impact on performance, or 20 GiB while still improving  the closed book baseline significantly. For comparison, the raw text of English Wikipedia requires approximately 24 GiB, thus making \modelname{} suitable for pre-computing compressed representations.

\end{minipage}
\hfill 
\begin{minipage}{0.37\textwidth}
\vspace{-1em}
    \centering
    \includegraphics[width=0.9\linewidth]{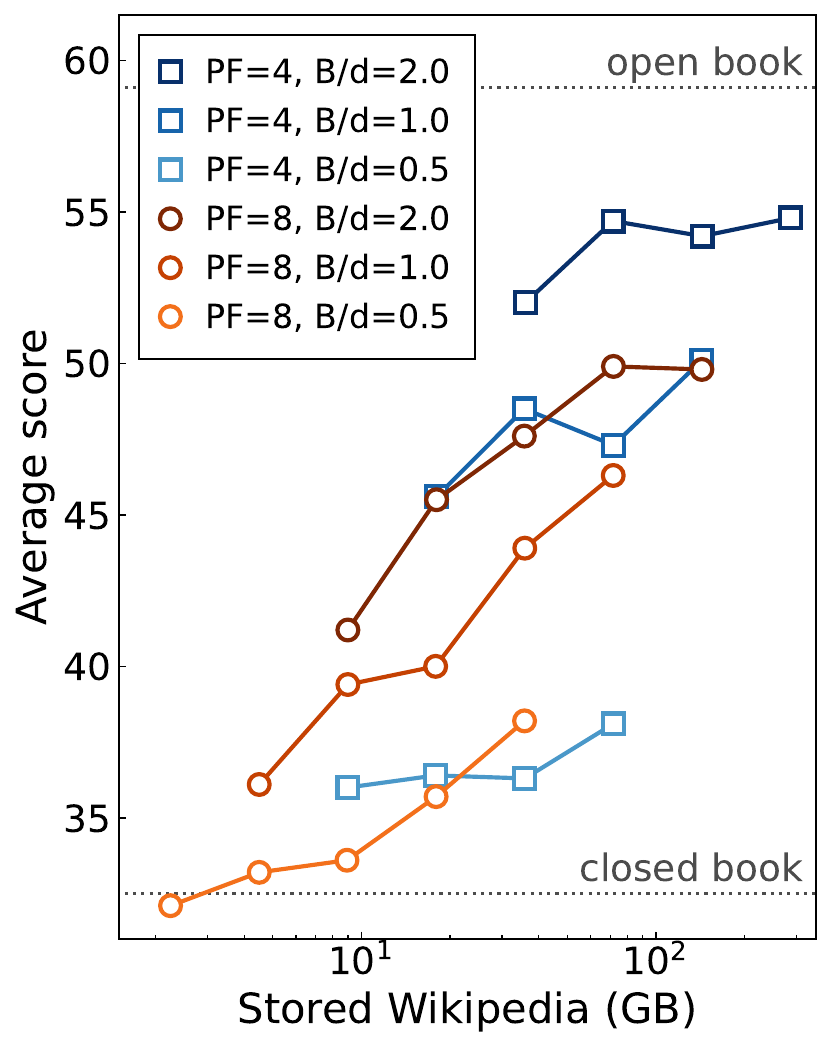}
    \captionof{figure}{Compression results with varying MLP dimensional bottlenecks and number of bits per dimension (B/d).}
    \label{fig:memory_bottleneck}
\end{minipage}
\footnotetext{\url{https://github.com/facebookresearch/faiss/wiki}}

\subsection{In-Domain and Out-of-Domain Analysis}

\begin{table}[h!]
\caption{
\textbf{Performance when adding HotpotQA and SQuAD train sets to the fine-tuning dataset.} Same setting and specifics as in Tab. 1.
}
\centering
\small
\setlength{\tabcolsep}{4pt}
\begin{tabular}{c l c c c c c c c c c}
\toprule
& Method & PF & Param.& \textbf{NQ} & \textbf{TQA} & \textbf{HQA} & \textbf{SQuAD}& \textbf{FLORES} & \textbf{CNN} & \textbf{Avg.}\\
\midrule
\multirow{5}{*}{\rotatebox{90}{\textit{Mistral 7B}}} 
& \textit{closed-book}\cellcolor{gray!10} &$\infty$\cellcolor{gray!10} &\cellcolor{gray!10} & $29.1$\cellcolor{gray!10} &$62.4$\cellcolor{gray!10} &$22.8$\cellcolor{gray!10} &$17.1$\cellcolor{gray!10} & \cellcolor{gray!10} & \cellcolor{gray!10} & \cellcolor{gray!10}\\
&\textit{open-book}\cellcolor{gray!10} &$1\times$\cellcolor{gray!10} &\cellcolor{gray!10} &$39.9$\cellcolor{gray!10} & $70.5$\cellcolor{gray!10} & $48.3$\cellcolor{gray!10} &$77.7$\cellcolor{gray!10} & $31.3$\cellcolor{gray!10} & $27.2$\cellcolor{gray!10} & $49.2$\cellcolor{gray!10}\\
& \modelname{$_4$}$^\otimes$\cellcolor{blue!8}    & $4\times$\cellcolor{blue!8}    & $3.0$B\cellcolor{blue!8} & $38.3$\cellcolor{blue!8}    & $\mathbf{68.9}$\cellcolor{blue!8}    & $\underline{60.5}$\cellcolor{blue!8}    & $\underline{77.0}$\cellcolor{blue!8}    & $29.9$\cellcolor{blue!8}    & $\mathbf{26.0}$\cellcolor{blue!8}    & $\underline{50.1}$\cellcolor{blue!8}    \\
& \modelname{$_4$}$^M$\cellcolor{blue!8}    & $4\times$\cellcolor{blue!8}    & $-$\cellcolor{blue!8}    &$\underline{38.4}$\cellcolor{blue!8}    & $67.9$\cellcolor{blue!8}    & $\mathbf{60.7}$\cellcolor{blue!8}    & $\mathbf{81.1}$\cellcolor{blue!8}    & $\underline{30.9}$\cellcolor{blue!8}    & $\underline{22.7}$\cellcolor{blue!8}    & $\mathbf{50.3}$\cellcolor{blue!8}    \\
& \modelname{$_8$}$^M$\cellcolor{blue!8}    & $8\times$\cellcolor{blue!8}    & $-$\cellcolor{blue!8}    & $\mathbf{39.0}$\cellcolor{blue!8}    & $67.0$\cellcolor{blue!8}    & $57.5$\cellcolor{blue!8}    & $74.8$\cellcolor{blue!8}    & $28.0$\cellcolor{blue!8}    & $20.8$\cellcolor{blue!8}    & $47.9$\cellcolor{blue!8}    \\
\bottomrule
\end{tabular}
\label{tab:ft_w_trainsets_short} 
\end{table}

\myparag{In-Domain Specialization.} In our primary evaluation, we deliberately excluded the training sets of the benchmark tasks to rigorously assess generalization. However, adapting \modelname{} to a specific domain remains a highly relevant use case for practitioners. In Tab.~\ref{tab:ft_w_trainsets_short}, we report the performance of models fine-tuned on our standard dataset augmented with the HotpotQA and SQuAD training sets. The results demonstrate that \modelname{} can be effectively specialized for targeted domains without requiring any modifications to the frozen decoder. Notably, the augmented models surpass the \textit{open-book} baseline on the injected domains (HotpotQA and SQuAD) and achieve a higher overall average, all while preserving high compression efficiency and maintaining competitive accuracy on the remaining tasks.

\begin{table}[h!]
\caption{\textbf{Out-of-domain evaluations.} Same setting as in Tab.~\ref{table:main_results}. Evaluated using the F1 score.}
\centering
\small
\setlength{\tabcolsep}{4pt}
\begin{tabular}{c ll c c c}
\toprule
& Methods & PF & Param. & \textbf{PubMedQA} & \textbf{BioASQ} \\
\midrule
\multirow{9}{*}{\rotatebox{90}{\textit{Mistral 7B decoder}}} 
& \textit{closed-book}\cellcolor{gray!10} &$\infty$\cellcolor{gray!10} &\cellcolor{gray!10} & $58.7$\cellcolor{gray!10} &$63.6$\cellcolor{gray!10}\\
&\textit{open-book}\cellcolor{gray!10} &$1\times$\cellcolor{gray!10} &\cellcolor{gray!10} &$84.4$\cellcolor{gray!10} & $77.5$\cellcolor{gray!10}\\
& ICAE-like$^\dagger$  &  $\sim 32$ & $7.2$B  & $63.6$ & $57.7$ \\
& LLMLingua2 &  $1.9\times$  & $0.6$B  & $74.4$ & $73.4$ \\
& PISCO-like$^\dagger$& $\sim 32$ & $7.2$B   & $62.4$ & $60.5$ \\
\cmidrule{2-6}
& \modelname{$_4$}$^\otimes$\cellcolor{blue!8}    & $4\times$\cellcolor{blue!8}    & $3.0$B  \cellcolor{blue!8} & $73.9$\cellcolor{blue!8}    &$72.0$\cellcolor{blue!8}  \\
&\modelname{$_4$}$^M$\cellcolor{blue!8}    & $4\times$\cellcolor{blue!8}    & $-$\cellcolor{blue!8}    & $\underline{76.6}$\cellcolor{blue!8}    &$\textbf{75.6}$\cellcolor{blue!8}  \\
&\modelname{$_8$}$^M$\cellcolor{blue!8}    &$8\times$\cellcolor{blue!8}    &$-$\cellcolor{blue!8}    & $\textbf{76.7}$\cellcolor{blue!8}    &$\underline{74.9}$\cellcolor{blue!8}  \\
\bottomrule
\end{tabular}
\label{table:results_bio_domain_short}
\end{table}

\myparag{Out-of-Domain Generalization.} To evaluate robustness in out-of-domain scenarios, we test our models on BioASQ \citep{bioasq} and PubMedQA \citep{jin2019pubmedqadatasetbiomedicalresearch}. Because the biomedical domain is entirely absent from our fine-tuning mixture, this setup provides a rigorous test of real-world generalization to specialized professional domains. As shown in Tab.~\ref{table:results_bio_domain_short}, baseline compressors (which, excluding LLMLingua2, share our standard fine-tuning data) struggle to adapt to these novel contexts. In contrast, \modelname{} achieves strong performance across all tested pooling factors.

\section{Limitations \& Conclusion}

While our compression method improves efficiency and generalizes well across domains, it remains highly dependent on the task formatting of the fine-tuning dataset. This reliance hinders its adaptability to novel tasks, notably limiting its effectiveness in long-context summarization scenarios. Furthermore, the method struggles to maintain performance at higher compression ratios and has thus far only been used to perform context extension on models with relatively small native context windows compared to current standards.

We introduce \modelname{}, a novel method to compute compressed text representations that can replace the raw text input in large language models. By reducing the context length, our method leads to faster prefilling and decoding stages, while leaving the target LLM unchanged. We show that a single encoder can be trained to work with multiple decoders, or even extended to new decoders with minimal adaptation. This opens the way towards \emph{universal compressed representations}. These representations can be used for various tasks (e.g., QA, summarization, translation...) and generalize well to new domains. We show that pretraining and fine-tuning are both critical for the success of our approach. In terms of architecture, pooling in the attention mechanism leads to strong results, while allowing a constant pooling factor for different sequence sizes, as opposed to memory tokens. Finally, using an MLP between the encoder and decoder allows our approach to compress representations further and learn a single encoder for multiple decoders.

\subsection*{Statement of Broader Impact}
While \modelname{} improves computational efficiency, users should exercise caution in privacy-sensitive domains such as healthcare or legal analysis. Although compressed representations are continuous vectors, they do not guarantee data confidentiality as they remain susceptible to inversion or interpretation by specialized models.

\bibliography{main}
\bibliographystyle{tmlr}
\newpage

\appendix
\section{Further experiments}
\label{sec:supplementary_experiments}

\subsection{Context Compression with benchmark train sets} 
To demonstrate the generalization ability of \modelname{}, we deliberately avoid fine-tuning on the training sets of evaluation benchmarks. However, we believe that this could be an interesting use case for users who wish to specialize \modelname{} in a given domain. In Tab.~\ref{tab:ft_w_trainsets}, we report results from models fine-tuned on our dataset augmented with the HotpotQA and SQuAD training sets. These results show that, even without altering decoders, \modelname{}s can specialize effectively on specific benchmarks. They outperform the \textit{open-book} baseline while maintaining efficiency gains and without harming performance on other tasks.

\begin{table}[h!]
\caption{
\textbf{Performance when adding HotpotQA and SQuAD train sets to the fine-tuning dataset.}
`PF': the token reduction factor (e.g., $4\times$) for fixed-ratio methods or the number of compressed tokens used, e.g. $\sim 16$, when this number is fixed; `Param.': number of parameters of the encoder; The superscript on \modelname{} indicates if the model is specifically trained for one decoder ($^M$ for Mistral or $^L$ for Llama) or both simultaneously ($^\otimes$). Best context compression results are in \textbf{bold}, second best are \underline{underlined}.
}
\centering
\small
\setlength{\tabcolsep}{4pt}
\begin{tabular}{c l c c c c c c c c c}
\toprule
& Method & PF & Param.& \textbf{NQ} & \textbf{TQA} & \textbf{HQA} & \textbf{SQuAD}& \textbf{FLORES} & \textbf{CNN} & \textbf{Avg.}\\
 \midrule

\multirow{5}{*}{\rotatebox{90}{\textit{Mistral 7B}}} & \textit{closed-book}\cellcolor{gray!10} &$\infty$\cellcolor{gray!10} &\cellcolor{gray!10} & $29.1$\cellcolor{gray!10} &$62.4$\cellcolor{gray!10} &$22.8$\cellcolor{gray!10} &$17.1$\cellcolor{gray!10} & \cellcolor{gray!10} & \cellcolor{gray!10} & \cellcolor{gray!10}\\
&\textit{open-book}\cellcolor{gray!10} &$1\times$\cellcolor{gray!10} &\cellcolor{gray!10} &$39.9$\cellcolor{gray!10} & $70.5$\cellcolor{gray!10} & $48.3$\cellcolor{gray!10} &$77.7$\cellcolor{gray!10} & $31.3$\cellcolor{gray!10} & $27.2$\cellcolor{gray!10} & $49.2$\cellcolor{gray!10}\\
&  \modelname{$_4$}$^\otimes$\cellcolor{blue!8}    &   $4\times$\cellcolor{blue!8}    &  $3.0$B\cellcolor{blue!8} &  $38.3$\cellcolor{blue!8}    &  $\mathbf{68.9}$\cellcolor{blue!8}    &  $\underline{60.5}$\cellcolor{blue!8}    &  $\underline{77.0}$\cellcolor{blue!8}    &  $29.9$\cellcolor{blue!8}    &  $\mathbf{26.0}$\cellcolor{blue!8}    &  $\underline{50.1}$\cellcolor{blue!8}    \\
&  \modelname{$_4$}$^M$\cellcolor{blue!8}    & $4\times$\cellcolor{blue!8}    &  $-$\cellcolor{blue!8}    &$\underline{38.4}$\cellcolor{blue!8}    &  $67.9$\cellcolor{blue!8}    &  $\mathbf{60.7}$\cellcolor{blue!8}    &   $\mathbf{81.1}$\cellcolor{blue!8}    &  $\underline{30.9}$\cellcolor{blue!8}    &  $\underline{22.7}$\cellcolor{blue!8}    &  $\mathbf{50.3}$\cellcolor{blue!8}    \\
& \modelname{$_8$}$^M$\cellcolor{blue!8}    & $8\times$\cellcolor{blue!8}    &  $-$\cellcolor{blue!8}    & $\mathbf{39.0}$\cellcolor{blue!8}    &  $67.0$\cellcolor{blue!8}    &  $57.5$\cellcolor{blue!8}    &  $74.8$\cellcolor{blue!8}    &  $28.0$\cellcolor{blue!8}    &  $20.8$\cellcolor{blue!8}    &  $47.9$\cellcolor{blue!8}    \\
\midrule
\multirow{5}{*}{\rotatebox{90}{\textit{Llama3.1 8B}}} &\textit{closed-book}\cellcolor{gray!10} &$\infty$\cellcolor{gray!10} & \cellcolor{gray!10} & $25.4$\cellcolor{gray!10} & $60.6$\cellcolor{gray!10} & $21.6$\cellcolor{gray!10} & $15.3$\cellcolor{gray!10} &    \cellcolor{gray!10} & \cellcolor{gray!10} & \cellcolor{gray!10} \\
&\textit{open-book}  \cellcolor{gray!10} &  $1\times$\cellcolor{gray!10} & \cellcolor{gray!10} &  $38.6$\cellcolor{gray!10} &  $67.1$\cellcolor{gray!10} &  $47.1$\cellcolor{gray!10} & $72.2$\cellcolor{gray!10} & $32.8$\cellcolor{gray!10} & $26.5$\cellcolor{gray!10} & $47.4$\cellcolor{gray!10} \\
& \modelname{$_4$}$^\otimes$\cellcolor{blue!8}    &   $4\times$\cellcolor{blue!8}    &  $3.0$B\cellcolor{blue!8} &  $38.5$\cellcolor{blue!8}    &  $\mathbf{69.7}$\cellcolor{blue!8}    &  $\underline{61.9}$\cellcolor{blue!8}    &  $\underline{78.6}$\cellcolor{blue!8}    &  $\mathbf{33.2}$\cellcolor{blue!8}    &  $\mathbf{26.0}$\cellcolor{blue!8}    &  $\underline{51.3}$\cellcolor{blue!8}    \\
& \modelname{$_4$}$^L$\cellcolor{blue!8}    & $4\times$\cellcolor{blue!8}    &  $-$\cellcolor{blue!8}    &  $\mathbf{40.7}$\cellcolor{blue!8}    & $\underline{68.5}$\cellcolor{blue!8}    &  $\mathbf{62.1}$\cellcolor{blue!8}    &  $\mathbf{82.3}$\cellcolor{blue!8}    &  $\mathbf{33.2}$\cellcolor{blue!8}    &   $\underline{22.2}$\cellcolor{blue!8}    &  $\mathbf{51.5}$\cellcolor{blue!8}    \\
&\modelname{$_8$}$^L$\cellcolor{blue!8}       &  $8\times$\cellcolor{blue!8}    &  $-$\cellcolor{blue!8}    &  $\underline{38.6}$\cellcolor{blue!8}    &  $67.6$\cellcolor{blue!8}    &  $59.1$\cellcolor{blue!8}    &  $76.0$\cellcolor{blue!8}    &  $\underline{30.1}$\cellcolor{blue!8}    &  $21.7$\cellcolor{blue!8}    &  $48.9$\cellcolor{blue!8}    \\

\bottomrule
\end{tabular}
\label{tab:ft_w_trainsets} 
\end{table}

 \subsection{Pooling factor generalization}
\label{sec:comp_rate}
\begin{minipage}[t]{0.59\textwidth}
\vspace{-13em}
Fig.~\ref{fig:cr_pt_ft_pairs} demonstrates that \modelname{} pretrained at a certain pooling factor can still be fine-tuned at another, sometimes even improving results on downstream tasks. This transfer works best as we use a smaller pooling factor than the one previously pretrained on. Notably, pretraining at $8\times$ seems to be particularly effective since we can then outperform any other pair on pooling factors of $4\times$ and  $8\times$. In contrast, models do not generalize well when fine-tuned to higher pooling factors. This capability greatly benefits the method since it enables us to reach better results at various pooling factors while pretraining fewer models. We train our best \modelname{$_4$} and \modelname{$_8$} from models pretrained using a pooling factor of $8$. When pooling too much performance degrades sharply as with a pooling factor of $32$ where the model has an averaged score of $33.1$.
\end{minipage}%
\hfill
\begin{minipage}[t]{0.38\textwidth}
    \centering
    \includegraphics[width=0.8\linewidth]{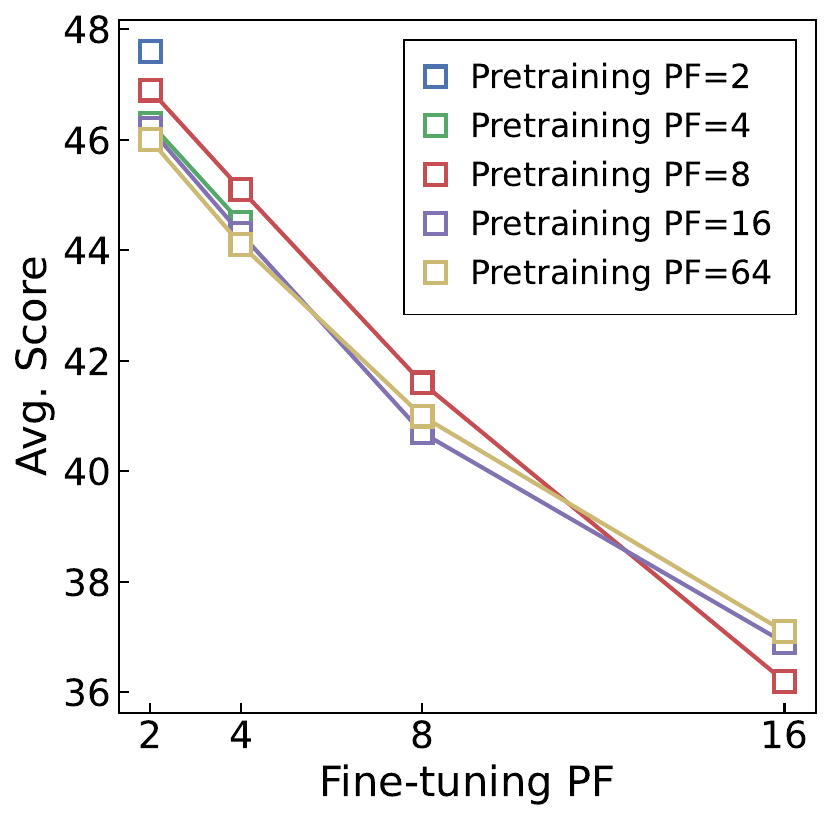}
    \captionof{figure}{Score for various pooling factors between pretraining and fine-tuning.}
    \label{fig:cr_pt_ft_pairs}
\end{minipage}

\subsection{Encoder-Decoder pairs} 
\label{sec:encoder_pairing}
Following the same recipe as described in \textsection~\ref{sec:method}, we can design \modelname{}\,/\,decoder pairs based on other backbone models. Tab.~\ref{tab:encoder_decoder_pairs} shows that Llama3.1 8B can also serve as an encoder backbone, displaying higher affinity with Llama decoders, similar to Llama3.2 3B. Adding the MLP projector is crucial to adapt to different decoders in the multi-decoder case as well as in the specific decoder one. 
Our experiments further reveal that the hidden state spaces of Llama3.1 8B and Mistral 7B are not fully disentangled: a single encoder can feed both decoders with the same compressed representations while outperforming the closed-book baseline. We attribute this compatibility to their similar architectures and training pipelines.

\begin{table}[H]
\caption{\textbf{Different backbones for an \modelname{$_4$} paired with different decoders}. The score is the average of the metrics in the Tab.~\ref{table:main_results}. For \underline{underlined} modules, the weights are the same for the two decoders; they are decoder-specific otherwise.}
\label{tab:encoder_decoder_pairs}
\centering
\setlength{\tabcolsep}{4pt}
\begin{tabular}{l c c c c c } \toprule
 & \multicolumn{5}{c}{Encoder}\\ 
\cmidrule{2-6}
Decoder & L8B  & L3B + MLP & \underline{L8B} & \underline{L8B} + MLP & \underline{L3B} + MLP\\
\midrule
\textit{Llama3.1 8B} & $46.0$ & $47.2$ & $32.4$ & $46.2$ & $45.5$\\
\textit{Mistral 7B} & $44.8$ & $45.1$ & $32.8$ & $41.8$ & $43.2$\\
\bottomrule
\end{tabular}
\end{table}

\subsection{Further baselines evaluations} 
\begin{table}[htbp]
\caption{\textbf{Further baselines evaluations.} Same setting and specifics as in Tab.~\ref{table:main_results}}
\centering
\small
\setlength{\tabcolsep}{4pt}
\begin{tabular}{c ll c c c c c c c c}
\toprule
& Methods & PF & Param. & \textbf{NQ} & \textbf{TQA} & \textbf{HQA} & \textbf{SQuAD}& \textbf{FLORES} & \textbf{CNN} & \textbf{Avg.}\\
 \midrule
\multirow{12}{*}{\rotatebox{90}{\textit{Mistral 7B decoder}}} & \textit{closed-book}\cellcolor{gray!10} &$\infty$\cellcolor{gray!10} &\cellcolor{gray!10} & $29.1$\cellcolor{gray!10} &$62.4$\cellcolor{gray!10} &$22.8$\cellcolor{gray!10} &$17.1$\cellcolor{gray!10} & \cellcolor{gray!10} & \cellcolor{gray!10} & \cellcolor{gray!10}\\
&\textit{open-book}\cellcolor{gray!10} &$1\times$\cellcolor{gray!10} &\cellcolor{gray!10} &$39.9$\cellcolor{gray!10} & $70.5$\cellcolor{gray!10} & $48.3$\cellcolor{gray!10} &$77.7$\cellcolor{gray!10} & $31.3$\cellcolor{gray!10} & $27.2$\cellcolor{gray!10} & $49.2$\cellcolor{gray!10}\\

 & ICAE-like$^\dagger$  &  $\sim 32$ & $7.2$B  & $36.5$ & $66.7$ & $24.3$ & $58.8$ & $28.3$ &$15.8$ & $38.4$\\

& xRAG-like$^\dagger$  &  $\sim 1$  & $7.1$B & $30.7$ & $65.2$ & $21.5$ & $23.9$ & $0.9$ &$14.6$&$26.1$\\
 & &  $\sim 8$  & $-$ & $28.7$ & $62.4$ & $21.8$ & $24.2$ & $9.0$ & $10.7$& $26.1$\\
& LLMLingua2 &  $1.9\times$  & $0.6$B & $\underline{38.8}$ & $69.0$ & $\underline{43.7}$ & $59.2$ &   $12.6$& $\underline{24.9}$& $41.4$\\
& &  $3.6\times$  & $-$ & $35.2$ & $66.6$ & $36.0$ & $42.0$ &  $4.3$& $22.1$& $34.4$\\
& PISCO-like$^\dagger$& $\sim 32$ & $7.2$B  & $34.7$ & $68.5$ & $24.9$ & $38.2$ & $\underline{33.6}$ & $19.2$& $36.5$\\
& &$4\times$ & $-$  & $36.6$ & $\underline{69.2}$ & $29.4$ & $48.1$ & $\mathbf{34.5}$ & $19.3$& $39.5$\\
\tightcmidrule{2-11}
& \modelname{$_4$}$^\otimes$\cellcolor{blue!8}    & $4\times$\cellcolor{blue!8}    & $3.0$B  \cellcolor{blue!8} & $38.2$\cellcolor{blue!8}    &$\mathbf{70.4}$\cellcolor{blue!8}    &$40.8$\cellcolor{blue!8}    &$\underline{69.2}$\cellcolor{blue!8}    & $29.5$\cellcolor{blue!8}    &$\mathbf{25.6}$\cellcolor{blue!8}    & $\underline{45.6}$\cellcolor{blue!8}   \\
&\modelname{$_4$}$^M$\cellcolor{blue!8}    & $4\times$\cellcolor{blue!8}    & $-$\cellcolor{blue!8}    & $\mathbf{39.0}$\cellcolor{blue!8}    &$68.9$\cellcolor{blue!8}    & $\mathbf{45.1}$\cellcolor{blue!8}    &  $\mathbf{71.1}$\cellcolor{blue!8}    & $31.0$\cellcolor{blue!8}    & $23.8$\cellcolor{blue!8}    & $\mathbf{46.5}$\cellcolor{blue!8}   \\
&\modelname{$_8$}$^M$\cellcolor{blue!8}    &$8\times$\cellcolor{blue!8}    &$-$\cellcolor{blue!8}    & $38.4$\cellcolor{blue!8}    & $67.9$\cellcolor{blue!8}    & $40.8$\cellcolor{blue!8}    & $62.0$\cellcolor{blue!8}    & $28.3$\cellcolor{blue!8}    & $22.9$\cellcolor{blue!8}    & $43.4$\cellcolor{blue!8}   \\
\midrule
\multirow{12}{*}{\rotatebox{90}{\textit{Llama3.1 8B decoder}}} &\textit{closed-book}\cellcolor{gray!10} &$\infty$\cellcolor{gray!10} & \cellcolor{gray!10} & $25.4$\cellcolor{gray!10} & $60.6$\cellcolor{gray!10} & $21.6$\cellcolor{gray!10} & $15.3$\cellcolor{gray!10} &    \cellcolor{gray!10} & \cellcolor{gray!10} & \cellcolor{gray!10} \\
&\textit{open-book}  \cellcolor{gray!10} &  $1\times$\cellcolor{gray!10} & \cellcolor{gray!10} &  $38.6$\cellcolor{gray!10} &  $67.1$\cellcolor{gray!10} &  $47.1$\cellcolor{gray!10} & $72.2$\cellcolor{gray!10} & $32.8$\cellcolor{gray!10} & $26.5$\cellcolor{gray!10} & $47.4$\cellcolor{gray!10} \\
& ICAE-like$^\dagger$  &  $\sim 32$& $7.6$B & $38.4$ &  $67.3$ & $20.5$ & $61.6$ & $31.3$ & $17.3$& $39.4$\\
& xRAG-like$^\dagger$ &  $\sim 1$  & $7.1$B & $28.0$ & $62.1$ & $21.7$ & $22.3$ &  $3.4$&$12.7$&$25.0$\\
&  &  $\sim 8$  & $-$ & $26.6$ & $61.1$ & $21.9$ & $22.3$ & $6.4$ & $12.6$&$25.2$\\
& LLMLingua2 &  $2.0\times$ & $0.6$B & $36.1$ & $66.3$ & $\underline{45.2}$ & $58.8$ &  $13.6$& $\underline{23.8}$ & $40.6$\\
& &  $3.9\times$  & $-$ & $34.2$ & $66.1$ & $37.2$ & $41.9$ &   $3.2$& $21.3$ & $34.0$\\
& PISCO-like$^\dagger$ & $\sim 32$ & $7.6$B & $35.1$ &  $69.4$ & $30.6$  & $40.5$ & $\underline{35.2}$ & $19.7$ & $38.4$\\
& & $4\times$& $-$ &  $37.9$ & $\underline{70.5}$ & $37.0$ & $57.2$ & $\mathbf{36.5}$& $20.7$ & $43.3$\\
 \tightcmidrule{2-11}
& \modelname{$_4$}$^\otimes$\cellcolor{blue!8}    & $4\times$\cellcolor{blue!8}    & $3.0$B\cellcolor{blue!8} & $\underline{39.6}$\cellcolor{blue!8}    & $\mathbf{70.8}$\cellcolor{blue!8}    & $43.6$\cellcolor{blue!8}    & $\underline{71.8}$\cellcolor{blue!8}    & $32.8$\cellcolor{blue!8}    & $\mathbf{26.1}$\cellcolor{blue!8}    & $\underline{47.5}$\cellcolor{blue!8}   \\
& \modelname{$_4$}$^L$\cellcolor{blue!8}    &  $4\times$\cellcolor{blue!8}    & $-$\cellcolor{blue!8}    & $\mathbf{39.7}$\cellcolor{blue!8}    & $70.1$\cellcolor{blue!8}    & $\mathbf{46.9}$\cellcolor{blue!8}    & $\mathbf{74.0}$\cellcolor{blue!8}    & $33.7$\cellcolor{blue!8}    & $23.7$\cellcolor{blue!8}    & $\mathbf{48.0}$\cellcolor{blue!8}   \\
& \modelname{$_8$}$^L$\cellcolor{blue!8}    &  $8\times$\cellcolor{blue!8}    & $-$\cellcolor{blue!8}    & $38.9$\cellcolor{blue!8}    & $69.0$\cellcolor{blue!8}    & $42.8$\cellcolor{blue!8}    & $66.0$\cellcolor{blue!8}    & $30.6$\cellcolor{blue!8}    & $22.8$\cellcolor{blue!8}    &  $45.0$\cellcolor{blue!8}   \\
\bottomrule
\end{tabular}
\label{table:main_results_full}
\end{table}

\subsubsection{ICAE further comparisons}
\label{subsub:icae_further_comparisons}
ICAE-style context compression relies on learned tokens (memory tokens) appended at the end of the context to compress. It produces for each encoded sequence a fixed number of compressed representations corresponding to the output of these memory tokens. In contrast, our method requires no additional learned tokens: the encoder directly pools context tokens, yielding a variable number of compressed representations that scales with input length. It enables consistent results across sequence lengths. Additionally, ICAE encoders mirror the architecture of their target decoder, while ARC-Encoders are trained starting from any decoder-only LLM (typically Llama 3.2 3B) with removed layers and bidirectional attention, making them noticeably smaller and more versatile.
ICAE-like model in Tab.~\ref{table:main_results} represents an ICAE context compression architecture under the same setting as our models training and evaluation. However, direct comparison is imperfect because ICAE’s pooling factor is not fixed: e.g. 32 memory tokens always produce 32 outputs, whereas ARC4-Encoder produces outputs 4x shorter than the input. To clarify this we have also trained and evaluated ICAE architectures on fixed-size chunks as described in Section 3.3.3 of the ICAE paper \citep{ge2024incontextautoencodercontextcompression}, so that they operate with a fixed pooling factor. We show downstream results of these models in Tab.~\ref{table:icae_morecomp} as well as the ICAE-like architecture with varying pooling factors across benchmarks. The downstream results show that our method outperforms ICAE-like models at matched pooling factors while requiring less than twice the encoding compute. For some benchmarks we still outperform ICAE-like models using ARC-Encoders with higher pooling factors.  

\begin{table}[htbp]
\caption{\textbf{Further baselines evaluations.} Same setting and specifics as in Tab.~\ref{table:main_results} with Mistral 7B decoder. Average pooling factors for each benchmark are indicated as subscripts of the reported scores. These variations occur when the number of memory tokens is fixed, and ICAE-like models are not trained or evaluated on fixed-size chunks, resulting in a pooling factor that depends on the context length.}
\centering
\footnotesize
\setlength{\tabcolsep}{4pt}
\begin{tabular}{ll c c c c c c c}
\toprule
Methods & PF & \textbf{NQ} & \textbf{TQA} & \textbf{HQA} & \textbf{SQuAD}& \textbf{FLORES} & \textbf{CNN} & \textbf{Avg.}\\
\midrule
\textit{closed-book}\cellcolor{gray!10} &$\infty$\cellcolor{gray!10} & $29.1$\cellcolor{gray!10} &$62.4$\cellcolor{gray!10} &$22.8$\cellcolor{gray!10} &$17.1$\cellcolor{gray!10} & \cellcolor{gray!10} & \cellcolor{gray!10} & \cellcolor{gray!10}\\
\textit{open-book}\cellcolor{gray!10} &$1\times$\cellcolor{gray!10} &$39.9$\cellcolor{gray!10} & $70.5$\cellcolor{gray!10} & $48.3$\cellcolor{gray!10} &$77.7$\cellcolor{gray!10} & $31.3$\cellcolor{gray!10} & $27.2$\cellcolor{gray!10} & $49.2$\cellcolor{gray!10}\\

 ICAE-like$^\dagger$  &  $\sim 32$ & $36.5_{\tiny(5\times)}$ & $66.7_{\tiny (5\times)}$ & $24.3_{\tiny (46\times)}$ & $58.8_{\tiny (6\times)}$ & $28.3_{\tiny (1\times)}$ &$15.8_{\tiny (32\times)}$ & $38.4$\\
 &  $4\times$ &  $36.4$ & $66.7$ & $23.8$ & $60.5.$ & $28.7$ & $18.6$ & $39.1$\\
 &  $\sim 16$ & $35.7_{\tiny (10\times)}$ & $66.7_{\tiny (9\times)}$ & $26.0_{\tiny (92\times)}$ & $51.0_{\tiny (12\times)}$ & $26.9_{\tiny (2\times)}$ & $14.3_{\tiny (64\times)}$ & $20.6$\\
 &  $8\times$ &  $34.8$ & $66.6$ & $8.9$ & $53.0$ & $26.7$ & $17.5$ & $34.6$\\
 &  $\sim 8$ & $34.9_{\tiny (19\times)}$ & $65.3_{\tiny (19\times)}$ & $25.3_{\tiny (185\times)}$ & $28.9_{\tiny (23\times)}$ & $18.5_{\tiny (4\times)}$ & $17.7_{\tiny (128\times)}$ & $18.1$\\
 &  $16\times$ & $33.7$ & $65.6$ & $0.1$ & $31.1$ & $19.0$ & $14.7$ & $27.4$\\
\midrule
\modelname{$_2$}$^M$\cellcolor{blue!8}    & $2\times$\cellcolor{blue!8} & $\mathbf{41.7}$\cellcolor{blue!8}    &$\mathbf{69.3}$\cellcolor{blue!8}    & $\mathbf{48.3}$\cellcolor{blue!8}    &  $\mathbf{76.7}$\cellcolor{blue!8}    & $\underline{30.4}$\cellcolor{blue!8}    & $18.9$\cellcolor{blue!8}    & $\mathbf{47.6}$\cellcolor{blue!8}   \\
\modelname{$_4$}$^M$\cellcolor{blue!8}    & $4\times$\cellcolor{blue!8}    &  $\underline{39.0}$\cellcolor{blue!8}    &$\underline{68.9}$\cellcolor{blue!8}    & $\underline{45.1}$\cellcolor{blue!8}    &  $\underline{71.1}$\cellcolor{blue!8}    & $\mathbf{31.0}$\cellcolor{blue!8}    & $\mathbf{23.8}$\cellcolor{blue!8}    & $\underline{46.5}$\cellcolor{blue!8}   \\
\modelname{$_8$}$^M$\cellcolor{blue!8}    &$8\times$\cellcolor{blue!8}   & $38.4$\cellcolor{blue!8}    & $67.9$\cellcolor{blue!8}    & $40.8$\cellcolor{blue!8}    & $62.0$\cellcolor{blue!8}    & $28.3$\cellcolor{blue!8}    & $\underline{22.9}$\cellcolor{blue!8}    & $43.4$\cellcolor{blue!8}   \\
\modelname{$_{16}$}$^M$\cellcolor{blue!8}    &$16\times$\cellcolor{blue!8}    & $35.4$\cellcolor{blue!8}    & $67.1$\cellcolor{blue!8}     & $31.8$\cellcolor{blue!8}     & $45.1$\cellcolor{blue!8}    & $22.7$\cellcolor{blue!8}     & $20.3$\cellcolor{blue!8}    & $37.1$\cellcolor{blue!8}\\   \modelname{$_{32}$}$^M$\cellcolor{blue!8}    &$32\times$\cellcolor{blue!8}    &$34.6$\cellcolor{blue!8}    & $65.8$\cellcolor{blue!8}     & $28.8$\cellcolor{blue!8}     & $34.8$\cellcolor{blue!8}    & $17.0$\cellcolor{blue!8}     & $17.8$\cellcolor{blue!8}    & $33.1$\cellcolor{blue!8}   \\
\bottomrule
\end{tabular}
\label{table:icae_morecomp}
\end{table}

\subsubsection{Generalization on out-of-domain professional datasets}
To assess the generalization ability of our fine-tuned models on strictly out-of-domain benchmarks, we evaluate the above models on BioASQ \citep{bioasq} and PubMedQA \citep{jin2019pubmedqadatasetbiomedicalresearch}. These two QA datasets focus on biological and biomedical question-answering. Both domains do not appear the fine-tuning datasets enabling to test out-of-domain generalization. It further highlights the strengths of the following models in specialized professional domains. Since all compressor models we compare against, except for LLMLingua2, were fine-tuned on the same dataset, they are expected to struggle with the out-of-domain nature of the provided contexts. As shown in Tab.~\ref{table:results_bio_domain}, our \modelname{} achieves the best performance even with a pooling factor of $8$, reaching results close to the upper-bound \textit{open-book} setting. These findings demonstrate the strong real-world applicability and robustness of our approach.

\begin{table}[h!]
\caption{\textbf{Out-of-domain evaluations.} Same setting and specifics as in Tab.~\ref{table:main_results}. Evaluated using F1 score.}
\centering
\small
\setlength{\tabcolsep}{4pt}
\begin{tabular}{c ll c c c}
\toprule
& Methods & PF & Param. & \textbf{PubMedQA} & \textbf{BioASQ} \\
 \midrule
\multirow{12}{*}{\rotatebox{90}{\textit{Mistral 7B decoder}}} & \textit{closed-book}\cellcolor{gray!10} &$\infty$\cellcolor{gray!10} &\cellcolor{gray!10} & $58.7$\cellcolor{gray!10} &$63.6$\cellcolor{gray!10}\\
&\textit{open-book}\cellcolor{gray!10} &$1\times$\cellcolor{gray!10} &\cellcolor{gray!10} &$84.4$\cellcolor{gray!10} & $77.5$\cellcolor{gray!10}\\
& ICAE-like$^\dagger$  &  $\sim 32$ & $7.2$B  & $63.6$ & $57.7$ \\
& LLMLingua2 &  $1.9\times$  & $0.6$B  & $74.4$ & $73.4$ \\
& &  $3.6\times$  & $-$  & $66.4$ & $69.7$ \\
& PISCO-like$^\dagger$& $\sim 32$ & $7.2$B   & $62.4$ & $60.5$ \\
& &$4\times$ & $-$   & $62.0$ & $66.0$ \\
\tightcmidrule{2-6}
& \modelname{$_4$}$^\otimes$\cellcolor{blue!8}    & $4\times$\cellcolor{blue!8}    & $3.0$B  \cellcolor{blue!8} & $73.9$\cellcolor{blue!8}    &$72.0$\cellcolor{blue!8}  \\
&\modelname{$_4$}$^M$\cellcolor{blue!8}    & $4\times$\cellcolor{blue!8}    & $-$\cellcolor{blue!8}    & $\underline{76.6}$\cellcolor{blue!8}    &$\textbf{75.6}$\cellcolor{blue!8}  \\
&\modelname{$_8$}$^M$\cellcolor{blue!8}    &$8\times$\cellcolor{blue!8}    &$-$\cellcolor{blue!8}    & $\textbf{76.7}$\cellcolor{blue!8}    &$\underline{74.9}$\cellcolor{blue!8}  \\
\midrule
\multirow{12}{*}{\rotatebox{90}{\textit{Llama3.1 8B decoder}}} &\textit{closed-book}\cellcolor{gray!10} &$\infty$\cellcolor{gray!10} & \cellcolor{gray!10} & $52.5$\cellcolor{gray!10} & $56.3$\cellcolor{gray!10} \\
&\textit{open-book}  \cellcolor{gray!10} &  $1\times$\cellcolor{gray!10} & \cellcolor{gray!10} &  $84.0$\cellcolor{gray!10} &  $77.2$\cellcolor{gray!10} \\
& ICAE-like$^\dagger$  &  $\sim 32$& $7.6$B  & $75.0$ & $70.9$ \\
& LLMLingua2 &  $2.0\times$ & $0.6$B  & $73.4$ & $75.1$ \\
& &  $4.0\times$  & $-$  & $62.7$ & $70.9$ \\
& PISCO-like$^\dagger$ & $\sim 32$ & $7.6$B  & $62.9$ & $67.8$ \\
& & $4\times$& $-$  & $68.4$ & $71.8$ \\
 \tightcmidrule{2-6}
& \modelname{$_4$}$^\otimes$\cellcolor{blue!8}    & $4\times$\cellcolor{blue!8}    & $3.0$B\cellcolor{blue!8} & $\textbf{81.1}$\cellcolor{blue!8}    & $\underline{76.3}$\cellcolor{blue!8}  \\
& \modelname{$_4$}$^L$\cellcolor{blue!8}    &  $4\times$\cellcolor{blue!8}    & $-$\cellcolor{blue!8}    & $\textbf{81.1}$\cellcolor{blue!8}    & $\textbf{77.1}$\cellcolor{blue!8}  \\
& \modelname{$_8$}$^L$\cellcolor{blue!8}    &  $8\times$\cellcolor{blue!8}    & $-$\cellcolor{blue!8}   & $\underline{80.2}$\cellcolor{blue!8}    & $75.5$\cellcolor{blue!8}  \\
\bottomrule
\end{tabular}
\label{table:results_bio_domain}
\end{table}

\subsection{Context length performance scaling}

\begin{figure*}[!ht]
    \centering
    \begin{subfigure}[t]{0.5\textwidth}
        \centering
        \includegraphics[width=0.9\linewidth]{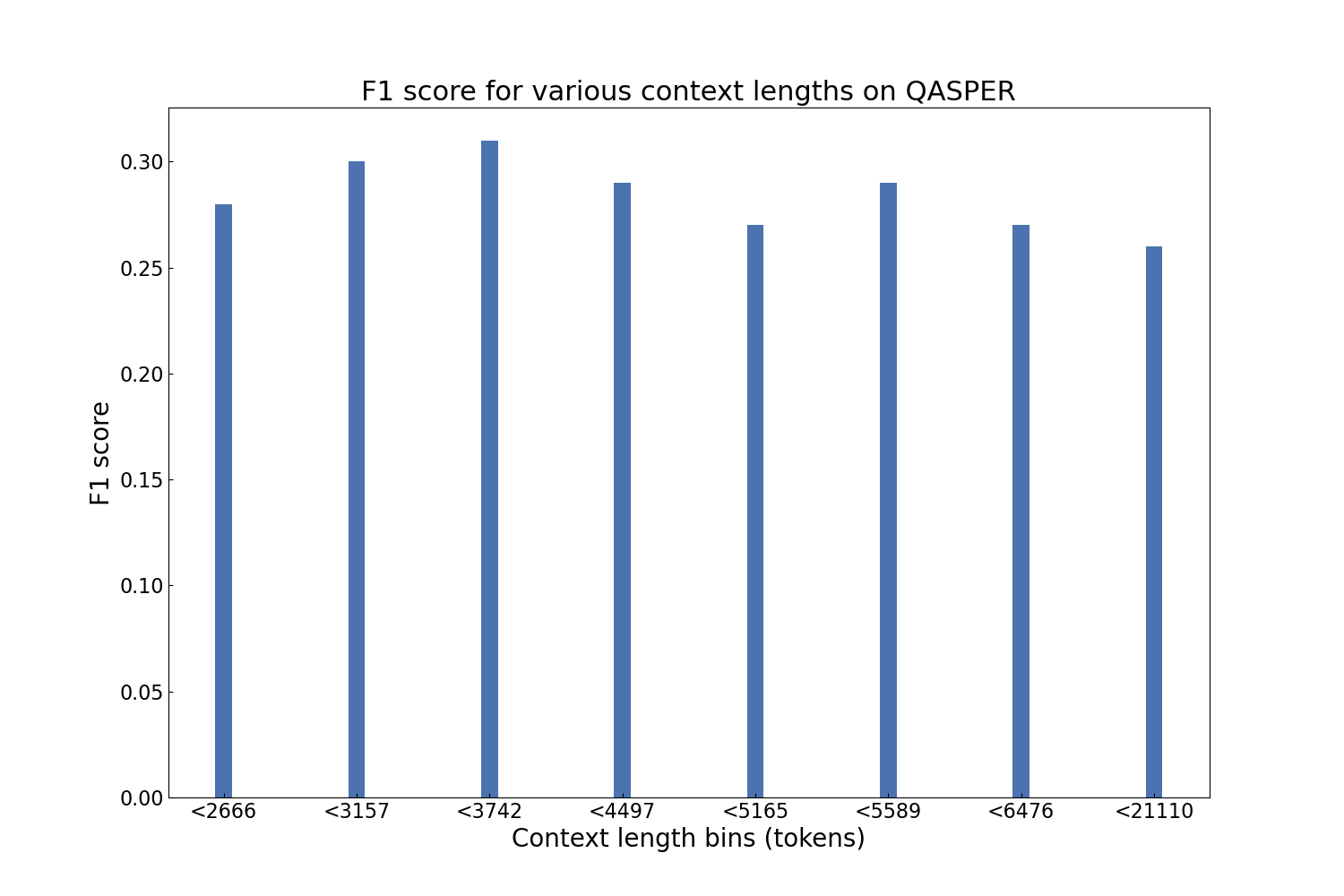}
        \caption{On QASPER (F1)}
    \end{subfigure}%
    \hfill
    \begin{subfigure}[t]{0.5\textwidth}
        \centering
        \includegraphics[width=0.9\linewidth]{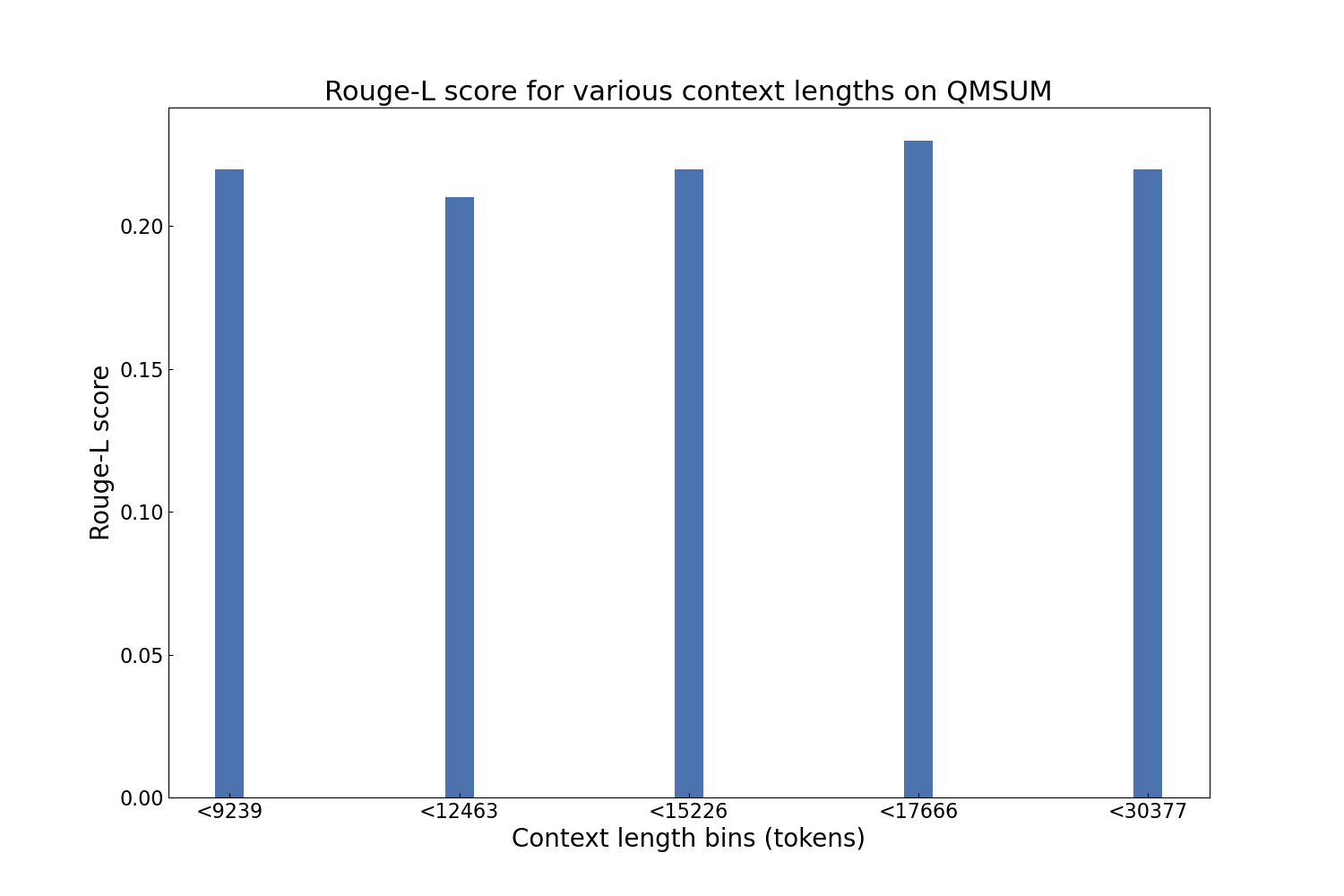}
        \caption{On QMSUM (Rouge-L)}
    \end{subfigure}
    \caption{\textbf{Performance scaling with context length}. On two benchmarks (a) QASPER and (b) QMSUM evaluation of our long-context ARC-Encoder paired with Llama2 7B Chat on subsets of the test dataset with similar context sizes.}
    \label{fig:context_length_scaling}
\end{figure*}

In Fig.~\ref{fig:context_length_scaling}, we show how performance scales with the context length in the long-context setting. We address this by splitting QMSUM and QASPER into bins containing approximately the same number of samples. Each bin groups examples whose contexts fall within a specific token-length range. We see that independently of the task (question answering for QASPER and summarization for QMSUM), ARC-Encoder performance remains largely consistent across different token-length ranges. We attribute this context length robustness to our fine-tuning which contains samples of varying lengths.

\subsection{Needle-in-Haystack analysis}
\label{subsec:niah}
\begin{figure*}[!ht]
    \centering
    \begin{subfigure}[t]{0.47\textwidth}
        \centering
        \includegraphics[width=\linewidth]{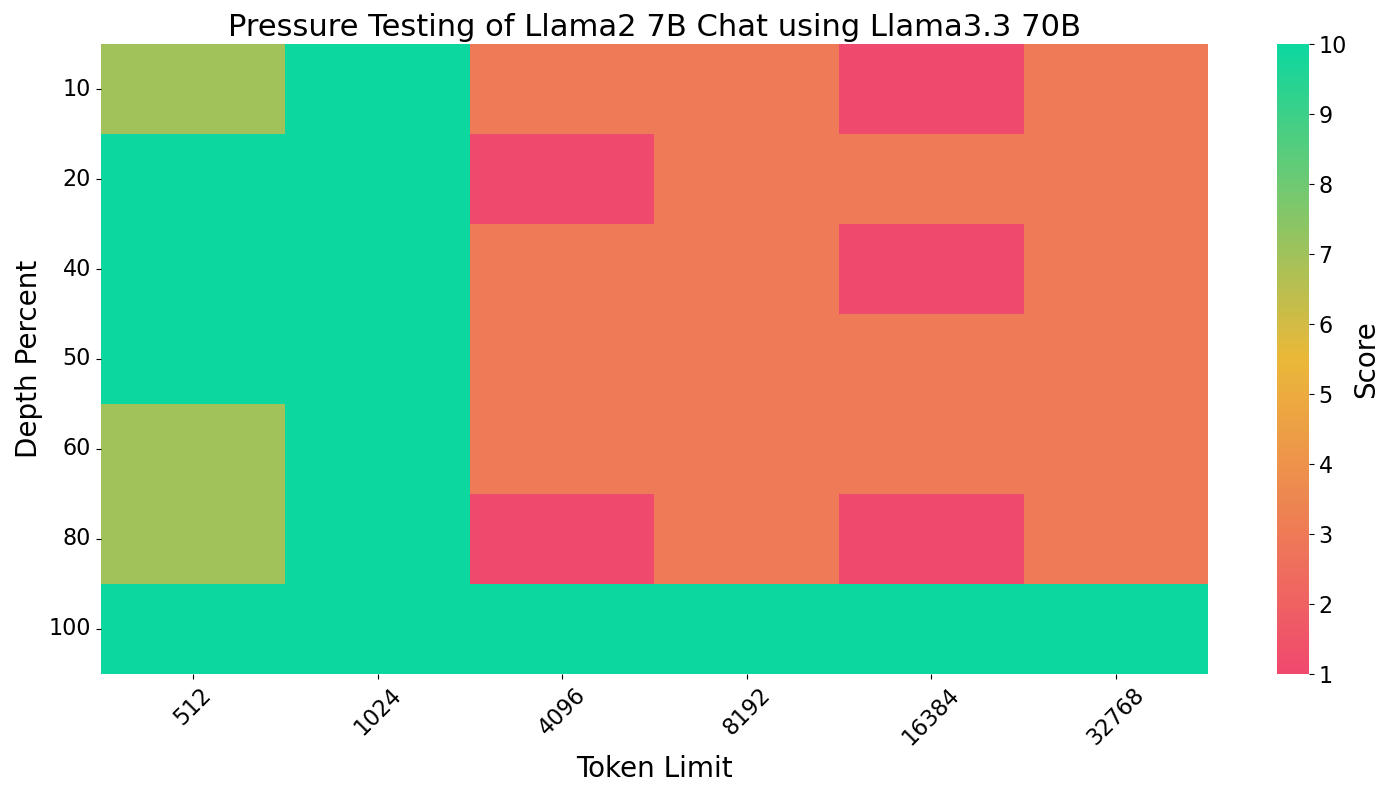}
    \end{subfigure}%
    \hfill
    \begin{subfigure}[t]{0.47\textwidth}
        \centering
        \includegraphics[width=\linewidth]{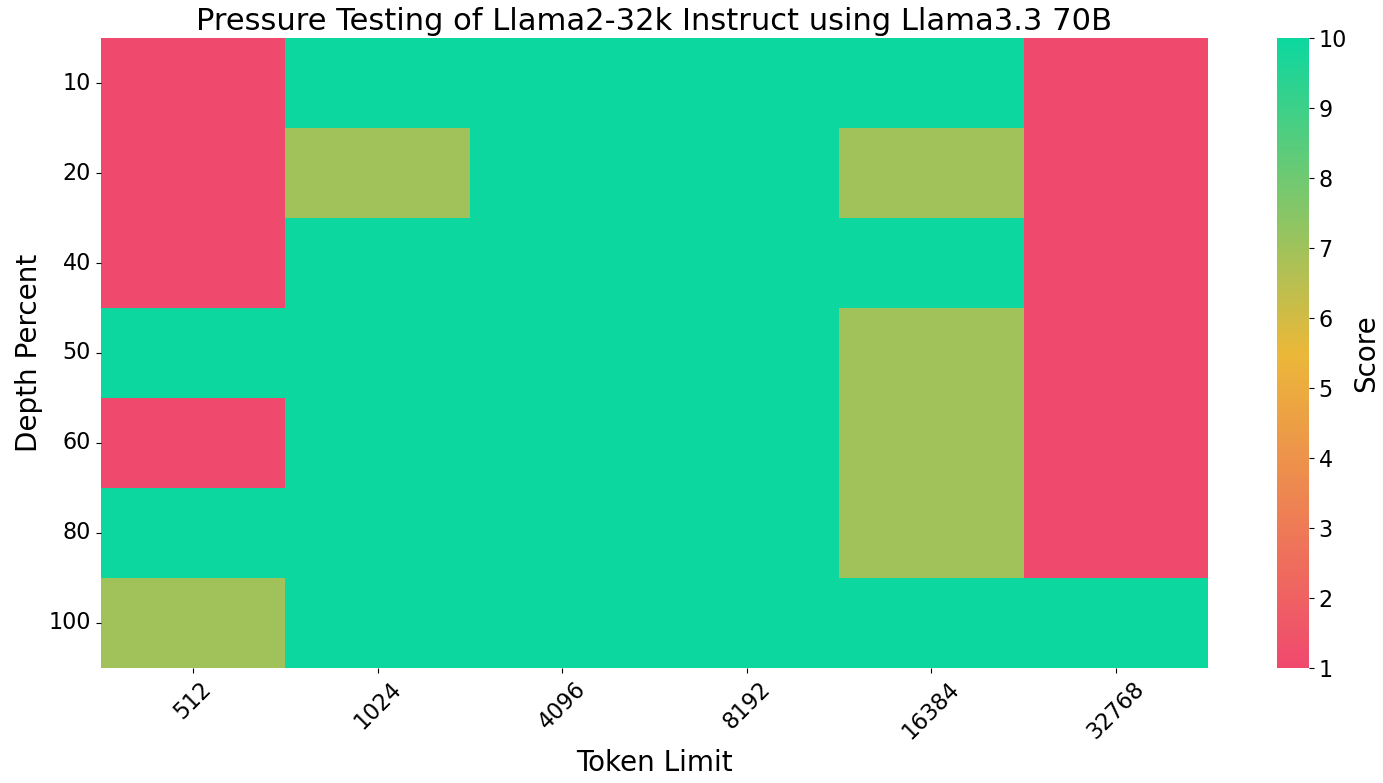}
    \end{subfigure}
        \centering
    \begin{subfigure}[t]{0.47\textwidth}
        \centering
        \includegraphics[width=\linewidth]{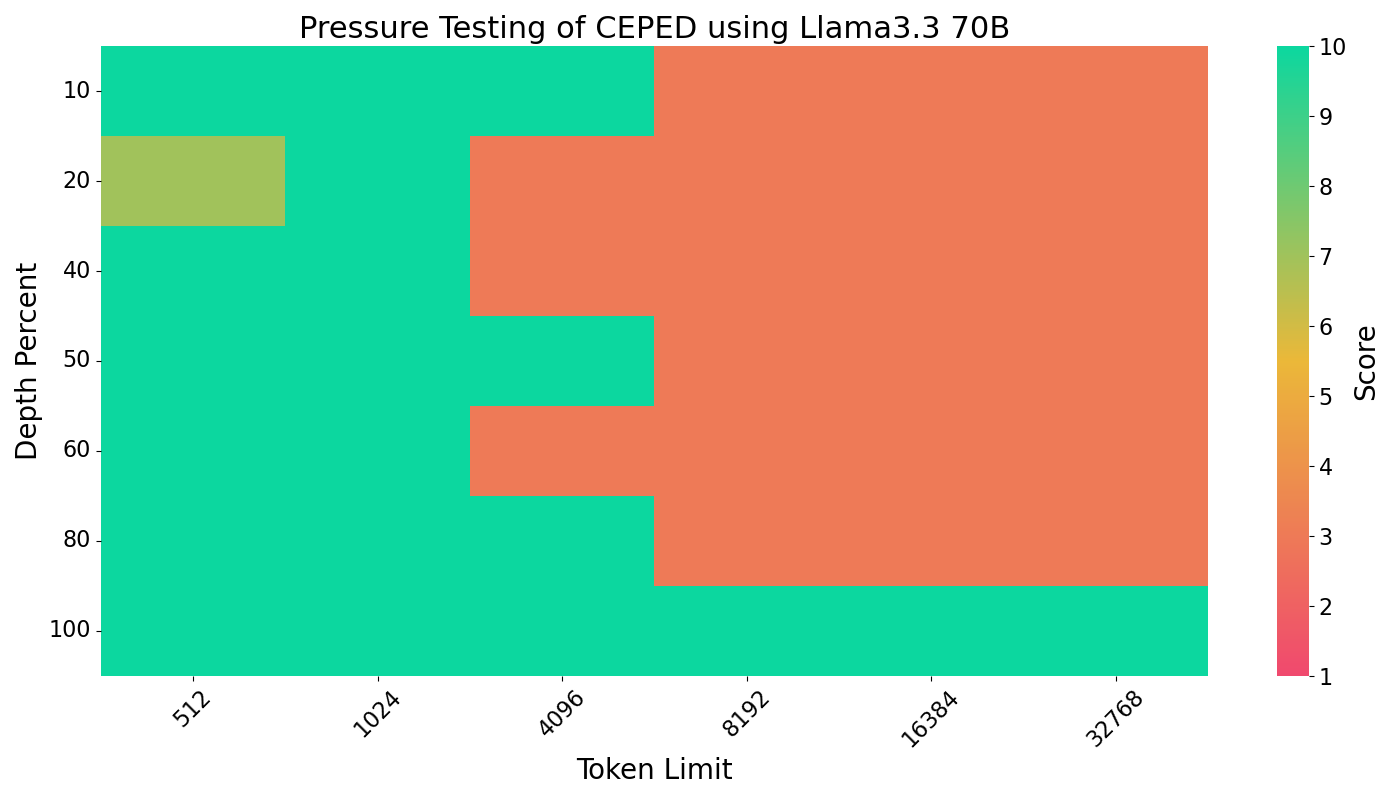}
    \end{subfigure}%
    \hfill
    \begin{subfigure}[t]{0.47\textwidth}
        \centering
        \includegraphics[width=\linewidth]{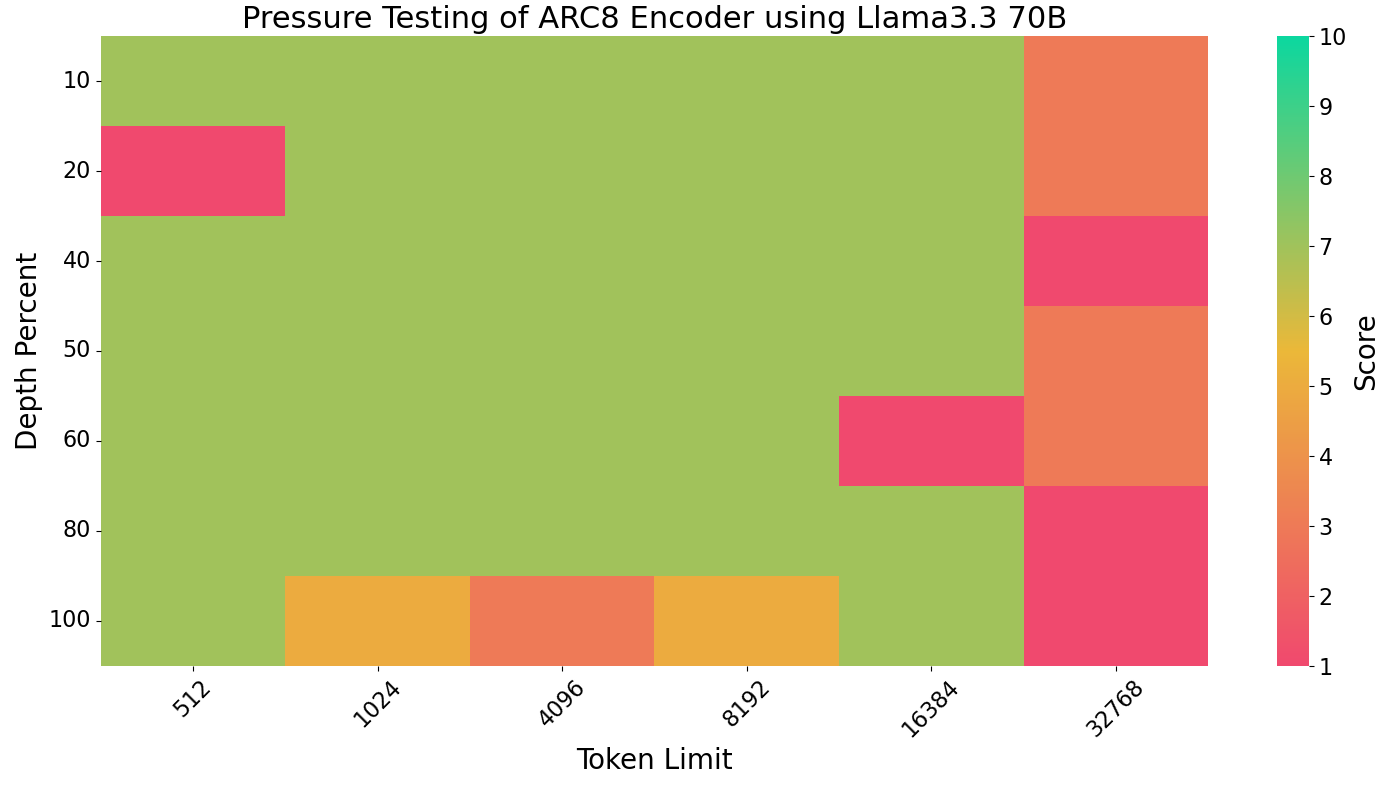}
    \end{subfigure}
    \caption{\textbf{Needle in the Haystack evaluation.} We test various models as in Tab.~\ref{tab:long_ctx}: Llama2 Chat (upper left), Llama2 32K fine-tuned for long-context (upper right), Llama2 with CEPED (bottom left) and our \modelname{$_8$} (bottom right).}
    \label{fig:niah}
\end{figure*}

Our method performs lossy context compression and our evaluation shows that this loss does not prevent the target LLM from extracting semantic information from the compressed representations for downstream tasks. To further explore this information loss, we evaluate fine-grained retrieval using the Needle-in-Haystack\citep{gkamradt2023LLMTest_NeedleInAHaystack} (NIAH) benchmark in Fig.~\ref{fig:niah}. In the following plots, we test ARC-Encoder (pooling factor 8) with Llama2 7B Chat in a long-context setting and compare it to other long-context-capable models, none of which were trained for this task. ARC-Encoder (bottom right) enables Llama2 7B Chat to retrieve precise information beyond its 4096-token window, and retrieval appears more consistent across varying context lengths and depths. Our model’s top score is 7 because it outputs ``Eat a sandwich and sit in Dolores Park on a sunny day'' instead of the exact needle given to the LLM as a judge ``The best thing to do in San Francisco is eat a sandwich and sit in Dolores Park on a sunny day.'' This demonstrates that ARC-Encoder enables fine-grained retrieval, but the model does not format the retrieved answer correctly.

\subsection{On targeting a smaller LLM}
\label{subsection:smaller_encoder}
\begin{figure*}[!ht]
    \centering
    \begin{subfigure}[t]{0.37\textwidth}
        \centering
        \includegraphics[width=\linewidth]{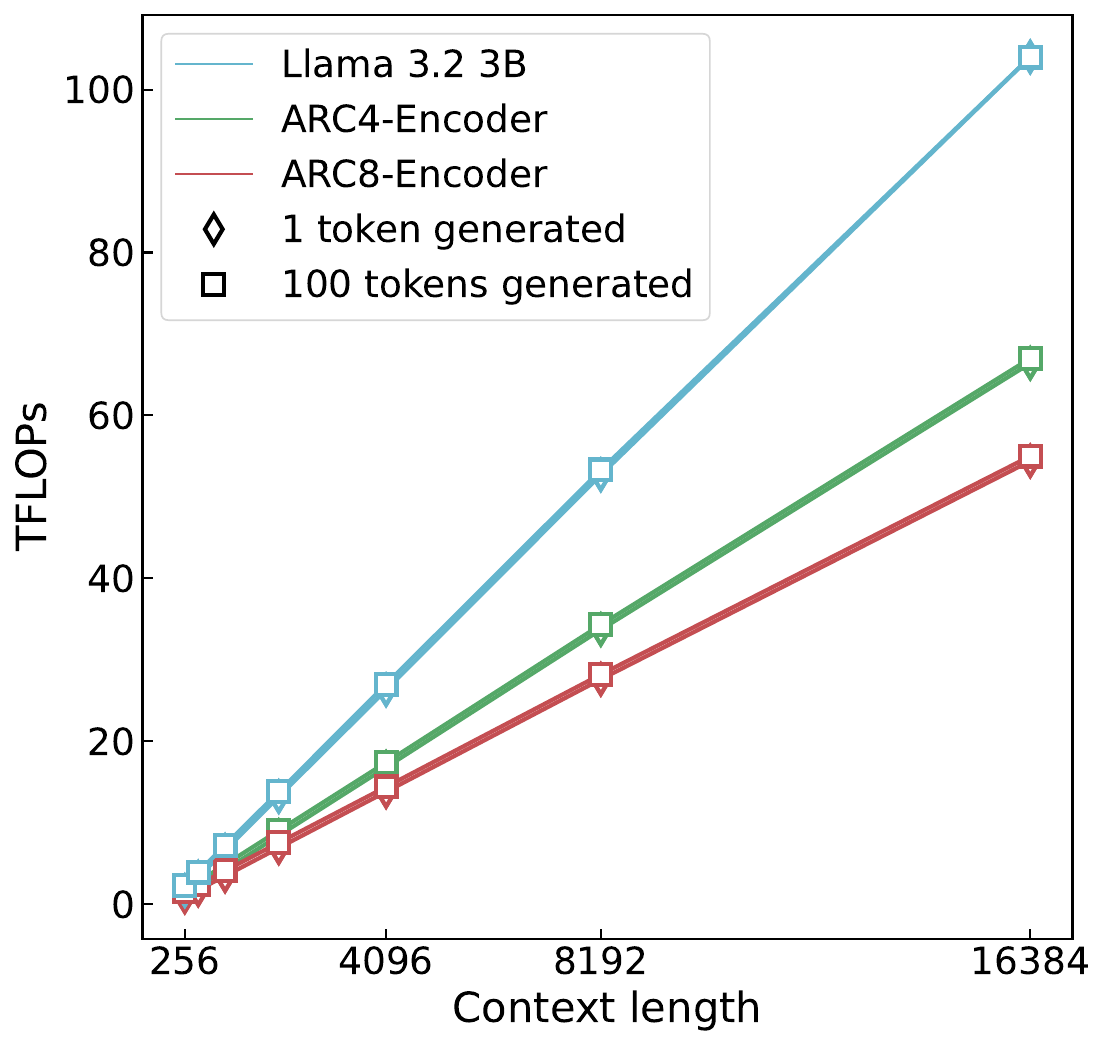}
        \caption{Number of TFLOPs}
    \end{subfigure}%
    \hfill
    \begin{subfigure}[t]{0.37\textwidth}
        \centering
        \includegraphics[width=\linewidth]{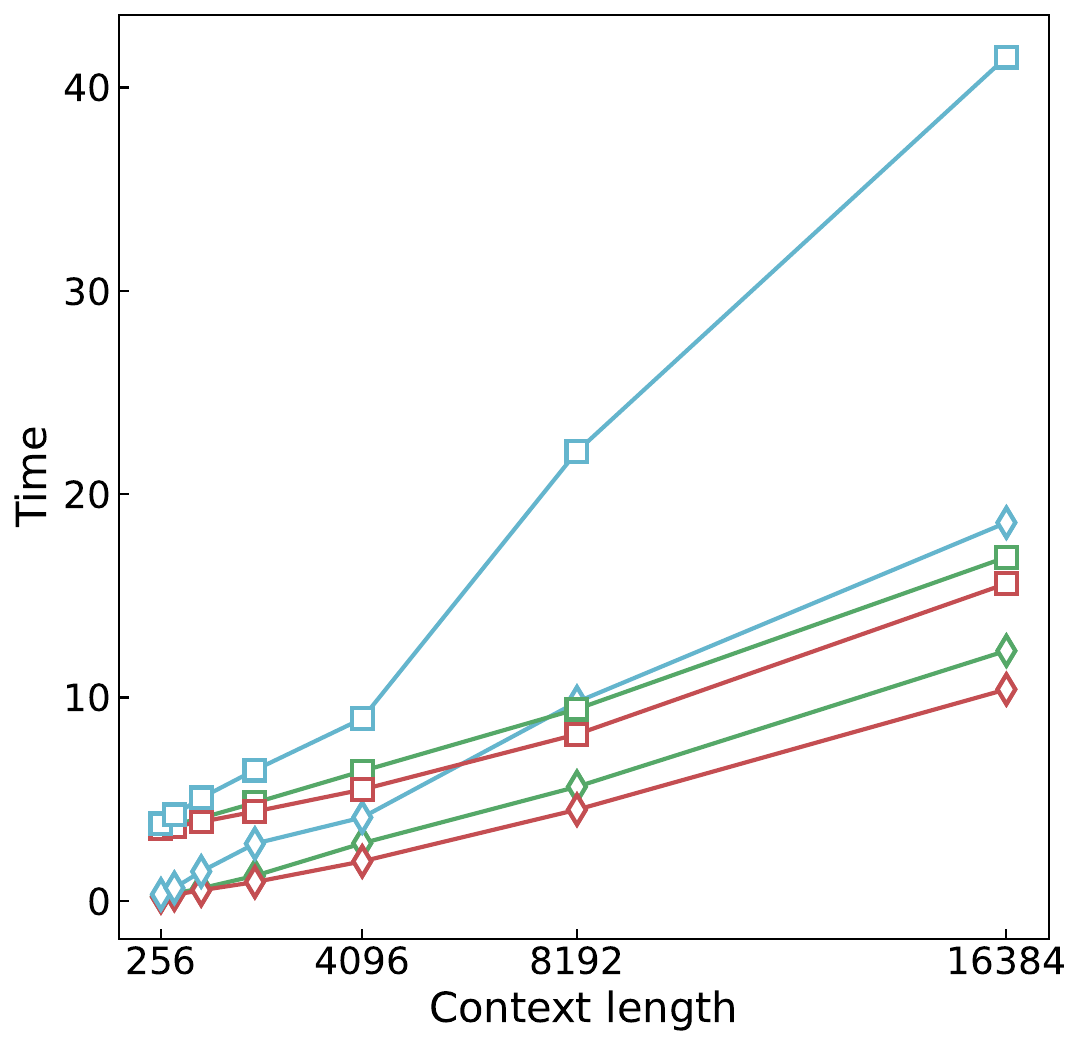}
        \caption{CUDA time (s)}
    \end{subfigure}
    \caption{\textbf{Measured computational costs for Llama 3.2 3B using ARC-Encoders with 1.8B parameters}. (a) Number of TFLOPs and (b) CUDA time in seconds for the continuation of a book from PG19 for various prompt lengths and numbers of tokens to generate on one NVIDIA H100.}
    \label{fig:flops_time_analysis_ACR_1.8B}
\end{figure*}

\begin{figure*}[!ht]
    \centering
    \begin{subfigure}[t]{0.37\textwidth}
        \centering
        \includegraphics[width=\linewidth]{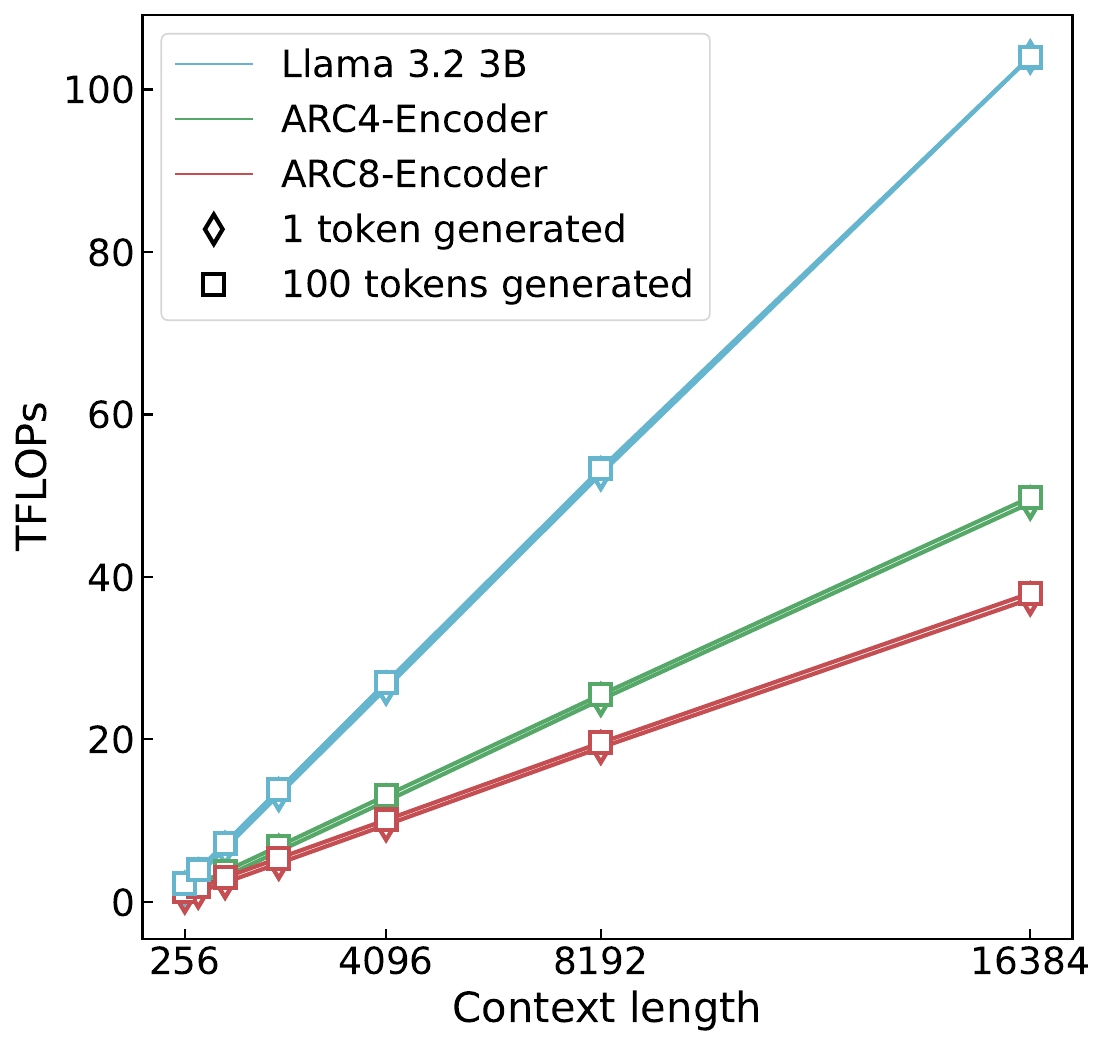}
        \caption{Number of TFLOPs}
    \end{subfigure}%
    \hfill
    \begin{subfigure}[t]{0.37\textwidth}
        \centering
        \includegraphics[width=\linewidth]{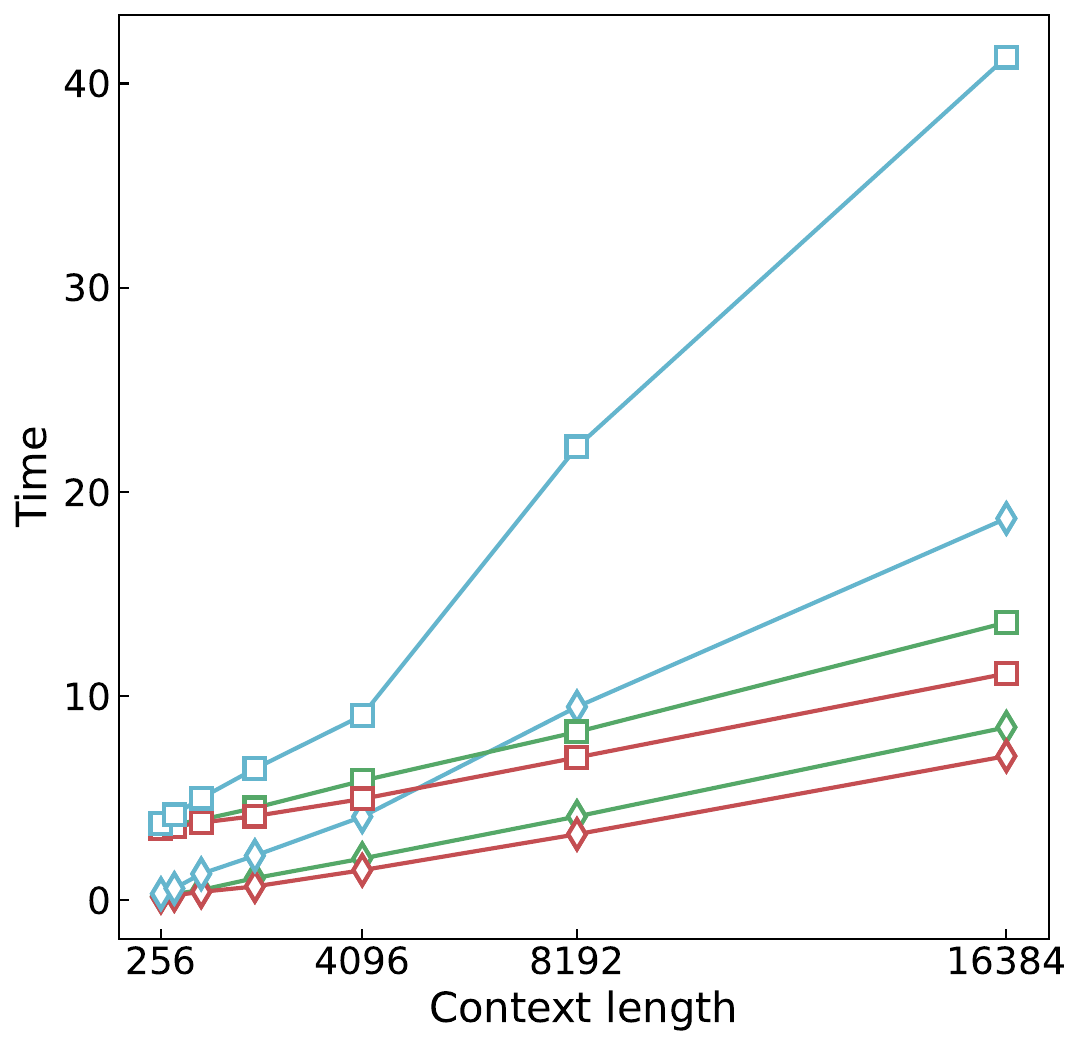}
        \caption{CUDA time (s)}
    \end{subfigure}
    \caption{\textbf{Measured computational costs for Llama 3.2 3B using ARC-Encoders with 1.1B parameters}. (a) Number of TFLOPs and (b) CUDA time in seconds for the continuation of a book from PG19 for various prompt lengths and numbers of tokens to generate on one NVIDIA H100.}
    \label{fig:flops_time_analysis_ACR_1.1B}
\end{figure*}

An ARC-Encoder with 3B parameters does not offer efficiency gains when paired with smaller LLMs. However, our encoder ablations show that we can truncate more layers from the backbone or even switch to a different backbone to reduce its size. We therefore train two additional ARC-Encoder variants: one using Llama 3.2 3B truncated by 14 layers (1.8B parameters) and one using Llama 3.2 1B truncated by 2 layers (1.1B parameters). When combined with a 3B LLM (Llama 3.2 3B), both variants achieve strong performance as shown in Tab.~\ref{table:small_models}, remaining close to the open-book setting, while also delivering computational benefits even at this scale, see Fig.~\ref{fig:flops_time_analysis_ACR_1.8B} and Fig.~\ref{fig:flops_time_analysis_ACR_1.1B}. Our method should be adapted depending on the targeted LLM to preserve an effective performance–efficiency trade-off.

\begin{table}[htbp]
\caption{\textbf{Smaller encoders paired with a smaller LLM.} Same setting and specifics as in Tab.~\ref{table:main_results} with Llama3.2 3B. We either use Llama3.2 3B with half of its layers truncated ($1.8$B parameters for the ARC-Encoder) or Llama3.1 1B truncated of 2 layers ($1.1$B parameters) as the backbone for an ARC-Encoder.}
\centering
\footnotesize
\setlength{\tabcolsep}{4pt}
\begin{tabular}{ll c c c c c c c c}
\toprule
Methods & PF & Param. & \textbf{NQ} & \textbf{TQA} & \textbf{HQA} & \textbf{SQuAD}& \textbf{FLORES} & \textbf{CNN} & \textbf{Avg.}\\
\midrule
\textit{closed-book}\cellcolor{gray!10} &$\infty$\cellcolor{gray!10} &  \cellcolor{gray!10} & $19.1$\cellcolor{gray!10} &$50.1$\cellcolor{gray!10} &$17.4$\cellcolor{gray!10} &$11.6$\cellcolor{gray!10} & \cellcolor{gray!10} & \cellcolor{gray!10} & \cellcolor{gray!10}\\
\textit{open-book}\cellcolor{gray!10} &$1\times$\cellcolor{gray!10}  & \cellcolor{gray!10} & $34.4$\cellcolor{gray!10} & $65.5$\cellcolor{gray!10} &$43.2$\cellcolor{gray!10} & $71.4$\cellcolor{gray!10} & $29.3$\cellcolor{gray!10} & $26.0$\cellcolor{gray!10}  & $45.0.$\cellcolor{gray!10}\\
\modelname{$_4$}$^L$\cellcolor{blue!8}    & $4\times$\cellcolor{blue!8} & $1.8$B\cellcolor{blue!8} & $37.13$\cellcolor{blue!8}    &$66.3$\cellcolor{blue!8}    & $40.3$\cellcolor{blue!8}    &  $65.0$\cellcolor{blue!8}    & $28.8$\cellcolor{blue!8}    & $21.0$\cellcolor{blue!8}    & $43.1$\cellcolor{blue!8}   \\
\modelname{$_8$}$^L$\cellcolor{blue!8}    & $8\times$\cellcolor{blue!8}    &   $-$\cellcolor{blue!8} &  $34.6$\cellcolor{blue!8}    &$65.5$\cellcolor{blue!8}    & $35.6$\cellcolor{blue!8}    &  $55.8$\cellcolor{blue!8}    & $24.7$\cellcolor{blue!8}    & $21.8$\cellcolor{blue!8}    & $39.7$\cellcolor{blue!8}   \\
\modelname{$_4$}$^L$\cellcolor{blue!8}    &$4\times$\cellcolor{blue!8}   &  $1.1$B\cellcolor{blue!8} &  $34.9$\cellcolor{blue!8}    &$65.9$\cellcolor{blue!8}    & $37.9$\cellcolor{blue!8}    &  $63.4$\cellcolor{blue!8}    & $27.6$\cellcolor{blue!8}    & $20.8$\cellcolor{blue!8}    & $41.8$\cellcolor{blue!8}   \\
\modelname{$_{8}$}$^L$\cellcolor{blue!8}    &$8\times$\cellcolor{blue!8}   &  $-$\cellcolor{blue!8} &  $33.0$\cellcolor{blue!8}    &$64.4$\cellcolor{blue!8}    & $33.2$\cellcolor{blue!8}    &  $52.4$\cellcolor{blue!8}    & $23.0$\cellcolor{blue!8}    & $20.5$\cellcolor{blue!8}    & $37.8$\cellcolor{blue!8}   \\
\bottomrule
\end{tabular}
\label{table:small_models}
\end{table}

\newpage
\section{Theoretical generation FLOPs}
\label{sec:th_flops}

Let us denote $s$ the number of tokens of the prompt sequence, $d$ the hidden dimension of the model, $n_{\text{layers}}$ its number of layers, $v$ the vocabulary size and $N$ the overall number of parameters of the decoder. We assume for simplicity that the decoder uses multi-head attention and that the hidden dimension is the same everywhere in the model. The theoretical complexity of prefilling steps, measured in floating point operations (FLOPs) and neglecting norms, is:
\begin{itemize}
    \item \textbf{Multi-head attention:}

\begin{align*}
    \textbf{Q, K, V projections}&=3\times s\times d^2\\
    \textbf{Attention scores}&=s^2\times d \\
    \textbf{$\times$ V}&=s^2\times d \\
    \textbf{Final projection}&= s\times d^2 (\text{depends})\\
    \textbf{Overall} &= 4s\times d^2 + 2 s^2\times d 
\end{align*}

    \item \textbf{Feedforward network} $\propto s\times d^2$
    \item \textbf{Output projection} $\propto s\times d\times v$\,.
\end{itemize}

In the case where $d \gg s$, as the number of parameters per layer is proportional to $d^2$, we have $d \propto \sqrt{\frac{N}{n_\text{layers}}}$, which leads to: 
\begin{equation}
\textbf{Total prefill FLOPs} \propto sN \quad \text{if } \sqrt{N} \gg s.
\end{equation}
Hence, using $\frac{s}{x}$ tokens for prefilling instead of $s$ leads to a relative FLOPs of $\frac{1}{x}$.

At the opposite, when processing very large prompts, i.e, $d \ll s$, the computational complexity reads:
\begin{equation}
\textbf{Total prefill FLOPs} \propto s^2\sqrt{N}. 
\end{equation}
If we use an \modelname{}, with a pooling factor of $x$, which has $n=p\times N$ parameters ($p<1$) with $p$ the relative size of the encoder vs. decoder, then the number of FLOPs to compress the prompt can be approximated to:
\begin{equation}
    \begin{cases}
    \propto sN\times (p+\frac{1}{x}) \quad &\text{if } d \gg s, \\
    \propto s^2\sqrt{N}(\sqrt{p} + \frac{1}{x^2}) \quad &\text{if } d \ll s. 
\end{cases}
\end{equation}
For instance, it is approximately $1.5\times$ smaller for  \modelname{$_4$} and $1.9\times$ smaller for  \modelname{$_8$} in the setting of the main table.

\section{Training details}
\label{training_details}

\subsection{Default setting}
\label{ablation_default}
Trainings are performed on $8 \times $H100 NVIDIA GPUs using PyTorch's FSDP framework\footnote{\url{https://docs.pytorch.org/docs/stable/fsdp.html}}. The ablations follow the parameters and architectural choices from our best \modelname{$_8$} encoders  \ref{table:main_results}, namely:
\begin{itemize}
\item \textbf{Encoder}: Llama3.2 3B truncated of the 2 last layers with every layer trained using a non-causal attention mask. When using Llama3.1 8B as encoder we truncate the 8 last layers. 
\item \textbf{Pooling}:  by averaging queries in the last transformer block of the encoder with a pooling factor of $8$.
\item \textbf{MLP projector}: 2 learned matrices without activation function, with dimensions sequence $3072 \rightarrow 2048 \rightarrow 4096$ if the encoder is Llama3.2 3B, $4096 \rightarrow 2048 \rightarrow 4096$ otherwise.
\item \textbf{Training}:\begin{itemize}
    \item Special tokens are added depending on the task.
    \item $20\%$ reconstruction for pretraining during $60$k steps with approximately $2$B tokens seen by the encoder and maximum $256$ tokens compressed and $256$ tokens to continue or reconstruct.
    \end{itemize}
    \item During the continuation task, maximum $256$ text tokens are prefixed to the compressed sequence to better align with the final few-shot evaluation.
\item \textbf{Fine-tuning}:\begin{itemize}
    \item \texttt{<Cont>} token is appended after each compressed sequence.
    \item Compressed representations are used as context and all the samples follow the format below (\ref{ft_template}) with more or less in-context examples as described in Tab. \ref{tab:finetuning_dataset}.
    \item Fine-tuning is performed with the same pooling factor as for the pretraining, unless stated otherwise. 
    \end{itemize}
\end{itemize}

\paragraph{Remarks.} Inserting text tokens before the compressed sequence in the continuation task introduces a significant compute overhead during pretraining. Yet, after fine-tuning with interleaved few-shot samples, it offers no gains in the specific-decoder setting. Standard continuation can substitute for the ``interleaved'' one in pretraining, as long as fine-tuning later interleaves compressed and normal tokens. For consistency, we keep interleaved continuation in reported results since it helps \modelname{} generalize better in the multi-decoder setting.

\begin{figure}[!ht]
    \centering
    \includegraphics[trim=0 0 0 0, clip, width=0.3\textwidth]{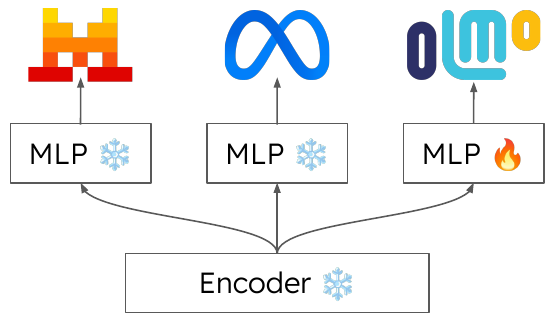}
    \captionof{figure}{Illustration of the extension of an \modelname{} to a new decoder by solely fine-tuning the MLP projector to the new decoder.}
    \label{fig:multi_decoder_adding}
\end{figure}

\subsection{Hyperparameters}

\begin{table}[h!]
\caption{Pretraining hyperparameters.}
\centering
\begin{tabular}{cc}
\toprule
Hyperparameters & Pretraining settings \\
\midrule
optimizer & AdamW \\
max lr encoder & $1\times10^{-5}$ \\
max lr special tokens & $1\times10^{-5}$ \\
max lr rate MLP & $5\times10^{-5}$ \\
lr scheduler type & 1 cycle policy \\
init lr & $1\times10^{-20}$ \\
final lr & $1\times10^{-10}$\\
warmup steps & 1000 \\
weight decay & 0.1 \\
batch size & 16 \\
gradient accumulation steps & None \\
GPUs & 8 \\
max tokens compressed at once  & 256\\
prefix text tokens & $\leq 256$ \\
max norm (gradient clipping) & 1.0 \\
mixed precision & Yes \\
number of steps & 60k for ablations\\
& 80k for multi-decoder \\
& 80k for \modelname{} (long context or not)\\
MLP init. & Kaiming Unif. leaky-ReLU slope of $\sqrt{5}$\\
Special tokens init. & Ones \\
\bottomrule
\end{tabular}
\label{tab:hyperparameters_pt}
\end{table}

\begin{table}[h!]
\caption{Fine-tuning hyperparameters. Not listed hyperparameters are identical to  pretraining ones.}
\centering
\begin{tabular}{ccc}
\toprule
Hyperparameters & Context Compression & Long Context \\
\midrule 
max lr encoder & $2\times10^{-6}$ & $-$\\
max lr special tokens & $2\times10^{-6}$ ($3\times10^{-5}$ for multi-decoder) & NA\\
max lr rate MLP & $3\times10^{-5}$ & $-$ \\
final lr & $1\times10^{-8}$& $-$\\
warmup steps & $50$ & $1000$\\
weight decay & $5\times10^{-2}$& $-$ \\
n. steps & $4$k ($8$k for multi-decoder) & $8$k \\
batch size & $8$ & $2$ \\
gradient accumulation steps & None & $4$ \\
max tokens compressed at once & $2048$ & $1024$ \\
interleaved examples & $\leq$ \# in-context examples ($M$) & NA\\
& $=M$ for Tabs. \ref{table:main_results}, \ref{table:result_add_decoder}, \ref{tab:ft_w_trainsets} and \ref{table:main_results_full}  & NA\\
contexts chunked to & No & $1024$ tokens \\ 
max contexts in parallel & $1$ & $31$ \\
\bottomrule
\end{tabular}

\label{tab:hyperparameters_ft}
\end{table}

\paragraph{Multi-Decoder specificities.} To add OLMo 7B to the existing multi-decoder \modelname{$_4$}, we fine-tune the MLP using $8$k steps with $100$ warmup steps and a maximum learning rate for the MLP of $10^{-4}$.

\subsection{Context Compression fine-tuning}
\label{fine_tuning_dataset}
To generate synthetic fine-tuning data, we use the vLLM\footnote{\url{https://docs.vllm.ai}} library for fast, efficient single-GPU inference. Gemma 3 27B \citep{gemmateam2025gemma3technicalreport} is then prompted to generate translations of Atlas Wikipedia passages (up to $3$ concatenated passages) in various languages that we split into two categories: 
\begin{itemize}
    \item[] 1) Spanish, French, German and Danish;
    \item[] 2)  Hindi, Russian, Swahili, Arabic, Turkish, Japanese, Finnish and Chinese (simplified).
\end{itemize}

We mix these generated translation datasets with supervised QA, summarization and reading comprehension datasets. For QA datasets, we retrieve the top-5 passages using NV-Embed \citep{lee2025nvembedimprovedtechniquestraining}, based on Wikipedia sequences\footnote{\url{https://huggingface.co/datasets/dmrau/kilt-128}} from KILT framework \citep{petroni2021kiltbenchmarkknowledgeintensive}.
To improve information retrieval in large compressed contexts, we concatenate retrieved passages when possible and compress them jointly. The number of concatenated passages is randomly sampled between $1$ and the maximum number of retrieved passages. This is particularly beneficial for HotpotQA and CNN, with minimal impact on other evaluation benchmarks.
For MS MARCO \citep{msmarco} we removed the samples without answers (\textit{no answer}). The complete list of the subsets that make up our final fine-tuning dataset is reported in Tab. \ref{tab:finetuning_dataset}, along with the proportions according to which these subsets are sampled.

\begin{table}[H]
\centering
\small
\begin{tabular}{l c c c} \toprule
& &\# in-context & Max. concat.\\
Subsets & Proportion & examples & passages\\
\midrule
\rowcolor{blue!10}
\multicolumn{4}{c}{Synth. translations} \\
Group 1 translations & $6\%$ & $5$ & \\
Group 2 translations  & $4\%$ & $5$ &\\
\rowcolor{blue!10}
\multicolumn{4}{c}{QA with retrieved context} \\
AdversarialQA \citep{Bartolo_2020} & $8\%$ & $5$ & $4$\\
FreebaseQA \citep{jiang-etal-2019-freebaseqa} & $27\%$ & $5$ & $4$\\
ASQA \citep{stelmakh2023asqafactoidquestionsmeet} & $1\%$ & $5$ & $4$\\
MS MARCO \citep{msmarco}& $9\%$ & $5$ & $4$\\
SciQ \citep{SciQ}& $3\%$ & $5$ & $4$\\
\rowcolor{blue!10}
\multicolumn{4}{c}{Reading comprehension} \\
DROP \citep{dua2019dropreadingcomprehensionbenchmark}&  $20\%$ & $5$ &\\
ParaSCI  \citep{dong-etal-2021-parasci}&  $12\%$ & $5$ &\\
\rowcolor{blue!10}
\multicolumn{4}{c}{Summarization} \\
DialogSum \citep{chen-etal-2021-dialogsum}&  $2.5\%$ & $3$ &\\
SAMSum \citep{gliwa-etal-2019-samsum}&  $2.5\%$ & $4$ &\\
WikiSum \citep{cohen-etal-2021-wikisum}&  $5\%$ & $5$ &\\
\bottomrule
\end{tabular}
\caption{Fine-tuning dataset for context compression} 
\label{tab:finetuning_dataset}
\end{table}

\begin{tcolorbox}[colback=blue!8,title=Few-shot compressed fine-tuning template]
    Document: \texttt{<TOKENS\_COMPRESSED>}\\
    Question: \texttt{<QUESTION>}\\
    Answer: \texttt{<ANSWER>}\\
    \\
    Document: \texttt{<TOKENS\_COMPRESSED>}\\
    Question: \texttt{<QUESTION>}\\
    Answer: \texttt{<ANSWER>}\\
    \hfill \vdots \\
    Document: \texttt{<TOKENS\_COMPRESSED>}\\
    Question: \texttt{<QUESTION>}\\
    Answer: \texttt{<ANSWER>} \quad \hfill (\textit{the loss is computed only on the last answer})\\
    \label{ft_template}
\end{tcolorbox}

\subsection{Long Context}
\subsubsection{Pretraining}
To pretrain \modelname{} when paired with an instruct decoder model, we continue to alternate between continuation and reconstruction tasks. We format each of the pretraining samples using the following template \ref{instruct_decoder_template}. 

\begin{tcolorbox}[colback=blue!8,title=Pretraining templates with Llama2 Chat decoder]
\textbf{Template:} \\
\texttt{<s>} [INST] Prefix+\texttt{<TOKENS\_COMPRESSED>}+Instruction [/INST] Suffix \texttt{</s>}\\
Reconstruction:
\begin{itemize}
\item Prefix = \texttt{"Text:\textbackslash n\textbackslash n"}
\item Instruction = \texttt{"\textbackslash n Replicate the input text."}
\item Suffix = \texttt{"Replicated text:\textbackslash n"}
\end{itemize}
Continuation:
\begin{itemize}
\item Prefix = \texttt{"Text:\textbackslash n\textbackslash n"}
\item Instruction = \texttt{"\textbackslash n Continue the previous text."}
\item Suffix = \texttt{"Text continuation:\textbackslash n..."}
\end{itemize}
\label{instruct_decoder_template}
\end{tcolorbox}

\subsubsection{Fine-tuning}

For this part, we synthesized QA, summarization and paraphrasing examples using the same procedure as in Appendix~\ref{fine_tuning_dataset}.

\textbf{QA generation.} We split books from PG-19 \citep{raecompressive2019} and arXiv papers from RedPajama \citep{weber2024redpajama} into paragraphs, randomly selecting 5 consecutive paragraphs. We then prompted Gemma3-27B to generate questions and gold answers using instructions such as:
\begin{itemize}
\item \textit{As a human instructor assessing students' comprehension of a scientific article, you craft a concise question that ideally requires a short phrase or sentence to answer. If the article lacks the necessary information, the answer should be `unanswerable'. For yes/no questions, reply with `yes', `no', or `unanswerable'. Then, supply the gold answer.}
\item \textit{You are given a story from a book. Your task is to create a question that can be answered in a short phrase or sentence. Then, provide the gold answer.}
\end{itemize}
The final context in the dataset is the entire book or paper. Additionally, we generate QA examples from Wikipedia by selecting one chunk from the Atlas Wikipedia dataset and appending up to 20 chunks from the same source before prompting the model.

\textbf{Summarization generation.} Using the same datasets, we split the texts into 10 groups of passages. Gemma3 27B is prompted to summarize each subsection; then, based on these summaries, it is prompted again to produce a higher-level summary, with prompt variations controlling the target length. For Atlas Wikipedia, we directly ask the model to produce a short summary from 10 consecutive chunks.
In both QA and summarization tasks, we truncate contexts longer than 500k characters and discard those shorter than 1k characters. 

\textbf{Paraphrase.} To mimic the questions asked in QM-Sum benchmarks, we prompt Gemma3 to reformulate passages of the text. For all tasks, we truncate contexts longer than 500k characters and discard those shorter than 1k characters. 

\begin{table}[H]
\centering
\small
\begin{tabular}{l c c c c } \toprule
Contexts & \# samples & Mean Ctx & Median Ctx & Mean answer\\
\midrule
\rowcolor{blue!10}
\multicolumn{5}{c}{Summarization} \\
From Atlas & $64000$& $6833$&$5593$  &$714$ \\
From PG-19 books & $64000$& $55048$& $34915$& $635$ \\
From ArXiv papers  & $40000$& $10896$& $4875$ &$834$ \\
\rowcolor{blue!10}
\multicolumn{5}{c}{Paraphrase} \\
From PG-19 books & $80000$& $31854$& $29272$& $665$ \\
From ArXiv papers  & $64000$& $5625$& $4260$ &$617$ \\
\rowcolor{blue!10}
\multicolumn{5}{c}{QA} \\
From Atlas & $80000$& $8065$ &$8325$ & $43$ \\
From PG-19 books &$80000$ & $317179$ &  $322942$ &$69$  \\
From ArXiv papers  & $40000$& $54526$ & $43386$ & $49$ \\
\bottomrule
\end{tabular}
\label{tab:longcontext_dataset}
\caption{\textbf{Fine-tuning dataset statistics for long-context understanding}. We report different statistics on the length in characters of the contexts (`Ctx') and the answers for each subset of fine-tuning samples.}
\end{table}
All samples are inserted in an instruction prompt depending on their task and context dataset using the same template as in \ref{instruct_decoder_template} with adapted \textit{Prefix}, \textit{Instruction} and \textit{Suffix}.

\section{Evaluation details}

\subsection{Baselines implementation}
\label{baselines_implem}

\paragraph{LLMLingua2 \citep{pan2024llmlingua2datadistillationefficient}:}
We used the open-source model \textit{microsoft/llmlingua-2-xlm-roberta-large-meetingbank} following the instructions from \href{https://github.com/microsoft/LLMLingua}{LLMLingua}.

\paragraph{xRAG \citep{cheng2024xragextremecontextcompression}:} We use the official codebase from \href{https://github.com/Hannibal046/xRAG}{xRAG} and extend it to support Llama3.1 8B as a decoder. Due to architectural similarities between Mistral 7B and Llama3.1 8B, only minor modifications are required. We first pretrain the MLP projector by closely following the instructions and data from the repository and the original paper, adapting it to base models by removing all chat templates. Next, we modify the fine-tuning dataset pre-processing to interleave compressed context in an ICL-style format, aligned with our fine-tuning template (see \ref{ft_template}), which closely matches the evaluation setup. To ensure consistency and avoid variability due to dataset size or quality, we use our own dataset for fine-tuning. 
We observe in Tabs. \ref{table:main_results} and \ref{table:main_results_full} that xRAG performs poorly on translations tasks (FLORES). After further investigations, we believe that compressing the full sequence into one vector leads to a loss of information that causes partial-only translations or hallucinations, as illustrated below.

\paragraph{ICAE \citep{ge2024incontextautoencodercontextcompression}:} For this re-implementation, we use our own codebase. We follow the hyperparameters and design choices described in the original paper, including the use of special tokens and alternating pretraining tasks. We set the language modeling task ratio to 0.5 and pretrain the encoder on our evaluated decoders (Llama3.1 8B and Mistral 7B) with our crawl dataset for $100$k steps (which is half the number of steps reported in the paper, but the training curve had already converged). Additionally, we adapt the fine-tuning template to match our evaluation format (see Appendix~\ref{ft_template}). To avoid redundancy with our ablation studies on pooling methods with memory tokens, we retain the fine-tuning dataset from \citet{ge2024incontextautoencodercontextcompression} (PwC), which was specifically synthesized for this purpose. However, due to the poor generalization on our evaluation benchmarks, we present ICAE-like models fine-tuned on our own dataset. 

\paragraph{PISCO \citep{louis2025piscoprettysimplecompression}:}
As with ICAE, we re-implemented PISCO using our own codebase, following the hyperparameters and design choices outlined in the original paper (e.g., LoRA applied to both encoder and decoder, encoder architecture, use of memory tokens). Since the official code is not publicly available, we referred to \citet{ge2024incontextautoencodercontextcompression} for implementation details not specified in the paper, such as the use of special tokens. To ensure consistency and avoid variability from dataset quality, we use our custom fine-tuning dataset but increased the number of fine-tuning steps to $8000$ to reach near $500$k samples which matches the number of training samples used in the PISCO paper. Furthermore, our custom dataset consists in PISCO fine-tuning dataset without the train sets of the evaluation datasets, with extra summarization datasets and synthesized translations data. While sequence-level distillation is key to avoiding reliance on gold labels, early experiments with silver labels showed that using gold labels enables a fairer comparison. Additionally, we train variants that process fixed-length input chunks of size $128$ tokens, enabling a fixed pooling factor and aligning the setup more closely with that of \citet{louis2025piscoprettysimplecompression}. 

\begin{tcolorbox}[colback=red!5,title=xRAG with Mistral 7B failures to translate English texts]
\textbf{To French:}
\begin{itemize}
    \item \texttt{Ground-truth:} ``Cette page est accessible facilement à partir d'une seule adresse Web, ce qui la rend facile à mémoriser et à écrire pour les étudiants qui ne savent pas utiliser un clavier et qui ont des problèmes d'orthographe.''
    \item \texttt{xRAG generation:}  ``Avec un seul nom de domaine, il est facile pour les utilisateurs d'accéder à l'information, ce qui est un avantage pour les étudiants.''
\end{itemize}
\textbf{To Spanish:}
\begin{itemize}
    \item \texttt{Ground-truth:} ``Son superiores a los servidores proxy por varios motivos: redirigen todo el tráfico de Internet y no únicamente los http.''
    \item \texttt{xRAG generation:}  ``Estos son más eficientes que los proxy, ya que no requieren que el usuario realice cambios en sus configuraciones de red.''
\end{itemize}
\textbf{To German:} 
\begin{itemize}
    \item \texttt{Ground-truth:} ``Vergessen Sie nicht die Extrakosten für weitere Visa, Abfluggebühren, Transportmittel an Land etc. für all die Orte außerhalb von Afrika mit einzuberechnen.''
    \item \texttt{xRAG generation:} ``Das ist zwar teurer als die Flüge, aber das Geld ist es wert, weil man damit nicht nur das Flugzeug spart, sondern auch die Kosten für die Übernachtung, die Verpflegung und die Reiseversicherung.''
\end{itemize}
\end{tcolorbox}

\subsection{Evaluation datasets}
\label{eval_data}

\paragraph{Context Compression.}
We evaluate our pipeline on question answering (QA) and reading comprehension tasks using the following benchmarks: HotpotQA \citep{yang2018hotpotqadatasetdiverseexplainable} (\textit{distractor} setting on the dev set, $7400$ samples) , Natural Questions \citep{47761} (NQ open dev set, $3605$ samples), TriviaQA \citep{joshi-etal-2017-triviaqa} (unfiltered nocontext validation set, $11308$ samples), SQuAD \citep{rajpurkar2016squad100000questionsmachine} ($10565$ samples). When ground-truth context is not provided, we retrieve the top-5 passages using NV-Embed \citep{lee2025nvembedimprovedtechniquestraining}, based on Wikipedia sequences from the Atlas framework \citep{izacard2022atlasfewshotlearningretrieval}, effectively simulating a RAG setup. The number of retrieved passages used for evaluation is specified for each benchmark. We report Exact Match (EM) as our primary evaluation metric, where answers are normalized and $\text{EM} = 1$ if all characters match exactly. We demonstrate summarization capabilities on the CNN-DailyMail dataset (a subset of $1000$ samples of the dev set), evaluating performance with the Rouge-L metric, as \citet{zhang2023benchmarkinglargelanguagemodels} noted that strong Rouge-L scores in this context are closely aligned with high human approval.

For translation tasks, we evaluate on the FLORES benchmark \citep{goyal2021flores101evaluationbenchmarklowresource} ($992$ samples), using BLEU scores computed with SacreBLEU\footnote{\url{https://github.com/mjpost/sacrebleu}}. BLEU scores are averaged over four translation directions: English to Danish, French, German, and Spanish. Models are prompted in a 5-shot setting using the following template, using compressed contexts for each example. Examples are sampled from the validation set and are fixed among all models. The reported pooling factor reflects the average per-context compression of tokens, not the ratio over the full prompt (including the textual prompt). It consists of dividing the number of tokens of the full document using the decoder tokenizer by the number of compressed tokens or the number of tokens of the compressed document in the hard compression case. 

\begin{tcolorbox}[colback=blue!8,title=Evaluation QA template]
\begin{itemize}
    \item \textbf{n examples for $n$-shot evaluation:}\\ 
    Document: \texttt{<TOKENS\_COMPRESSED>}\\
    Question: \texttt{<QUESTION>}\\
    Answer: \texttt{<ANSWER>}\\
    \\
    Document: \texttt{<TOKENS\_COMPRESSED>}\\
    Question: \texttt{<QUESTION>}\\
    Answer: \texttt{<ANSWER>}\\
    \hfill \vdots \\
    \item \textbf{the final question}\\
    Document: \texttt{<TOKENS\_COMPRESSED>}\\
    Question: \texttt{<QUESTION>}\\
    Answer: \\
\end{itemize}
\label{eval_templates}
\end{tcolorbox}

\begin{tcolorbox}[colback=blue!8,title=Evaluation translation template]
\begin{itemize}
    \item \textbf{n examples for $n$-shot evaluation:}\\ 
    Document: \texttt{<TOKENS\_COMPRESSED>}\\
    Question: Translate the previous document into \texttt{<LANGUAGE>}.\\
    Answer: \texttt{<ANSWER>}\\
    \hfill \vdots \\
    \item \textbf{the final question}\\
    Document: \texttt{<TOKENS\_COMPRESSED>}\\
    Question: Translate the previous document into \texttt{<LANGUAGE>}.\\
    Answer: \\
\end{itemize}
\end{tcolorbox}

\paragraph{Long Context.} For long-context understanding, we report results on NarrativeQA (NQA), QASPER (Qspr), GovReport (GvRp), and QM-Sum validation datasets from ZeroSCROLLS \citep{shaham2023zeroscrollszeroshotbenchmarklong} benchmark, a suite of zero-shot long-context understanding tasks that emphasize instruction-following capabilities. We evaluate on the full validation dataset which consists in respectively $3461$, $1726$, $973$ and $272$ samples. Specifically, we adopt the task formats and instructions as used in \citet{yen2024longcontextlanguagemodelingparallel}\footnote{\url{https://github.com/princeton-nlp/CEPE}}.

\end{document}